\pdfoutput=1
\documentclass[letterpaper]{article} 
\usepackage{times}  
\usepackage{helvet}  
\usepackage{courier}  
\usepackage[hyphens]{url}  
\usepackage[dvipdfmx]{graphicx} 
\usepackage[margin=1.5in]{geometry}
\usepackage[dvipsnames]{xcolor}
\usepackage{tikz}
\usepackage{tikzscale}
\usepackage{enumitem}
\usepackage{natbib}
\usetikzlibrary{arrows.meta}
\usetikzlibrary{decorations.pathreplacing}
\usepackage{natbib}  
\usepackage{caption} 
\usepackage{blindtext}
\usepackage{algorithm}
\usepackage{algorithmic}
 
\usepackage{microtype}
\usepackage{subfigure}
\usepackage{booktabs} 

\usepackage{amsmath}
\usepackage{amssymb}
\usepackage{mathtools}
\usepackage{amsthm}

\usepackage[capitalize,noabbrev]{cleveref}

\usepackage{enumitem}
\usepackage{natbib}  
\usepackage{caption} 
\usepackage{blindtext}
\usepackage{algorithm}
\usepackage{algorithmic}

\usepackage{microtype}

\usepackage{subfigure}
\usepackage{booktabs} 

\usepackage[capitalize,noabbrev]{cleveref}

\usepackage{multicol}
\usepackage[textsize=tiny]{todonotes}

\theoremstyle{plain}
\newtheorem{theorem}{Theorem}
\newtheorem{proposition}[theorem]{Proposition}
\newtheorem{lemma}{Lemma}

\newtheorem{definition}{Definition}

\theoremstyle{remark}

\newcommand{\algthrelong}{\texttt{Linear Threshold Greedy}\xspace}
\newcommand{\algthre}{\texttt{LinTG}\xspace}
\newcommand{\alglong}{\texttt{Linear Greedy}\xspace}
\newcommand{\alg}{\texttt{LG}\xspace}

\newcommand{\alglongstat}{\texttt{Linear Bandit with Static Strategy}\xspace}

\newcommand{\algstat}{\texttt{LBSS}\xspace}

\newcommand{\mR}{\mathbb{R}}

\newcommand{\vect}[1]{\textbf{#1}}

\newcommand{\appendixtitle}[1]{
	\begin{center}
		\LARGE \bf #1
	\end{center}
}
\newcommand{\Ha}{H_{l',a}^{(l)}}
\usepackage{multicol}
\usepackage[textsize=tiny]{todonotes}

\usepackage{authblk}
\title{Linear Submodular Maximization with Bandit Feedback}
\author[1]{Wenjing Chen}
\author[1]{Victoria G. Crawford}
\affil[1]{Department of Computer Science \& Engineering, Texas A\&M University}
\date{}                     
\begin{document}

\maketitle
\begin{abstract}
Submodular optimization with bandit feedback has recently been studied in a variety of contexts. In a number of real-world applications such as diversified recommender systems and data summarization, the submodular function exhibits additional linear structure. We consider developing approximation algorithms for the maximization of a submodular objective function $f:2^U\to\mathbb{R}_{\geq 0}$, where $f=\sum_{i=1}^dw_iF_{i}$. It is assumed that we have value oracle access to the functions $F_i$, but the coefficients $w_i$ are unknown, and $f$ can only be accessed via noisy queries. We develop algorithms for this setting inspired by adaptive allocation algorithms in the best-arm identification for linear bandit, with approximation guarantees arbitrarily close to the setting where we have value oracle access to $f$. Finally, we empirically demonstrate that our algorithms make vast improvements in terms of sample efficiency compared to algorithms that do not exploit the linear structure of $f$ on instances of move recommendation.


\end{abstract}
\section{Introduction}
Submodular set functions satisfy a diminishing returns property that arises in many optimization problems in machine learning such as those involving coverage functions \citep{bateni2017almost}, data summarization \citep{LinB11,tschiatschek2014learning}, facility location ~\citep{LindgrenWD16}, viral marketing in social networks \citep{kempe2003maximizing}, and recommendation systems~\citep{LeskovecKGFVG07,ElAriniG11}. A set function $f:2^U\to\mathbb{R}$ is submodular if for every $X\subseteq Y\subseteq U$ and for every $x\in U\setminus Y$, we have $f(X\cup\{x\})-f(X)\ge f(Y\cup\{x\})-f(Y)$. In addition, $f$ is often monotone, i.e., $f(Y)\ge f(X)$ for every $X\subset Y$. Among the most widely considered formulations is the NP-hard monotone submodular maximization problem with a cardinality constraint \cite{nemhauser1978analysis}: Given budget $\kappa$, find $\arg\max_{|S|\leq \kappa}f(S)$. The setting of most existing work on SM is that we have value oracle access to $f$, and algorithms may query any subset $X\subset U$ and the value of $f(X)$ is returned.

However, in many applications it is not realistic to assume that we have value oracle access to $f$, but rather that we can only query $f$ approximately or subject to some noise \citep{kempe2003maximizing,singla2016noisy,horel2016maximization,hassidim2017robust,parambath2018saga,peng2021dynamic}
Within this broad area of noisy access to $f$, one important line of work consider the development of algorithms for the special case that $f$ has additional linear structure \citep{yue2011linear,yu2016linear,chen2017interactive,hiranandani2020cascading}. In particular, the setting and problem is described below.
\begin{definition}[Linear Submodular Maximization with a Cardinality Constraint (SM)]
\label{def:prob}
      Let $U$ be a universe of elements of size $n$. Suppose we have $d$ monotone and submodular functions denoted as $F_i:2^U\rightarrow \mathbb{R}$ for $i=1,2,...,d$. The objective function $f:2^U\rightarrow \mathbb{R}$ is a weighted sum of the $d$ submodular functions, i.e. $f(S)=\sum_{i=1}^dw_iF_i(S)=\vect{F}^T(S)\vect{w}$,
where $\vect{F}(S)=(F_1(S),F_2(S),...,F_d(S))^T$, $\vect{w}=(w_1,w_2,...,w_d)^T$, and $w_i\geq 0$ for any $i\in[d]$. SM is then defined to be, given budget $\kappa$, find $\arg\max_{|S|\leq \kappa}f(S)$.
\end{definition}
Notice that $f$ is also monotone and submodular, since it is the linear sum of monotone submodular functions.
It is assumed that we have value oracle access to each of the basis functions $F_i$, and we can query them on any subset $S\subseteq U$, but the weight vector $\vect{w}=(w_1,w_2,...,w_d)^T$ is unknown. We do not have oracle access to $f$, but we may make i.i.d. R-sub-Gaussian noisy queries to $f$ where the expectation is the true value of $f$. An important application of this setting is recommendation systems, which we will more thoroughly describe in Section \ref{sect:prelim}. These assumptions on the structure of $f$ and noise fall into the well-studied linear bandit model \citep{soare2014best,xu2018fully}, as we will explain in more detail in Section \ref{sect:prelim}. The focus of existing work on SM in this setting is integrating linear bandit strategies and algorithms for SM together in order to select a subset $X\subseteq U$ over a series of iterations that minimizes regret (the online regret minimization setting) \citep{yue2011linear,yu2016linear,chen2017interactive,hiranandani2020cascading}. Existing works emphasize the online regret minimization setting, wherein the agent strives to select a solution set with high expected function values while exploring new sets.

In the online regret minimization setting, a solution set is selected at each iteration and some corresponding noisy feedback from $f$ is returned. It is assumed that there is a penalty to choosing a bad set, e.g. a set with low $f$ value is recommended to a user, and so a balance between exploration (sampling in order to better estimate $\vect{w}$) and exploitation (choosing sets at each iteration, e.g. to recommend to a user, that are believed to be of high value) is used in order to select sets. On the other hand, an alternative to the regret minimization setting is that of the pure exploration linear bandit \citep{soare2014best,xu2018fully}, where as few noisy samples as possible are taken in orders to arrive at a good solution quickly. We depart from existing literature on linear submodular bandit in that we do not use the online regret minimization setting, but instead make use of ideas from pure exploration linear bandit. More specifically, our goal is to provide a solution set with a PAC-style theoretical guarantee \cite{even2002pac,kalyanakrishnan2012pac}, where the bandit algorithm identifies an $\epsilon$-optimal arm with probability at least $1-\delta$, in as few samples as possible. This type of problem has been previously studied in the submodular optimization literature by \citeauthor{singla2016noisy} in the general case where $f$ does not have linear structure. But the linear structure of $f$ allows us to develop algorithms that make significantly fewer noisy samples (as will be demonstrated in Section \ref{sec:exp}), because we are able to \textit{learn} the unknown vector $\vect w$ as more and more noisy samples are made.

In particular, the contributions of our paper are as follows: (i) We propose and analyze the algorithm \alglong (\alg), which is inspired by the standard greedy algorithm, but proceeds as $\kappa$ rounds of best-arm-identification problems in the linear bandit setting. We prove that the exact function value of the output solution set of \alg $S$ is arbitrarily close to the optimal approximation guarantee of $1-1/e$ for cardinality constrained submodular maximization \citep{nemhauser1978analysis}. In addition, \alg chooses an optimal sample allocation strategy, and we prove a bound on the total number of noisy samples required by \alg. (ii) We next propose and analyse the algorthm \algthrelong (\algthre) which is based on the threshold greedy algorithm of \cite{badanidiyuru2014fast}, but we propose a new linear bandit strategy in a novel setting that is distinct from best-arm-identification. \algthre is more efficient compared to \alg from an algorithmic perspective, but additionally the linear bandit strategy itself is more efficient in several ways. Similar to \alg, \algthre is proven to return a solution set $S$ that is arbitrarily close to $1-1/e$. (iii) Finally, we compare our algorithms to several alternative approaches to this problem on instances of movie recommendation. We find that our algorithms make significant improvements in terms of sample complexity.

\subsection{Related Work}
There has been a great line of work on noisy submodular maximization  \cite{horel2016maximization,hassidim2017submodular,crawford2019submodular}. Among these works, the most related setting assumes the noisy feedback is i.i.d $R$-sub-Gaussian, which is also referred to as bandit feedback \cite{singla2016noisy,karimi2017stochastic}. The problem is also referred to as submodular bandit, which is a specific type of combinatorial multi-armed bandit problem with the reward function assumed to be submodular \cite{chen2017interactive,takemori2020submodular,hiranandani2020cascading}. The main focus of these works are regret minimization. Among the works studying submodular bandit, the linear submodular bandit is a specific problem with the same objective function as us \cite{yue2011linear,takemori2020submodular,yu2016linear}, i.e., the objective function $f$ can be expressed by the linear sum of known submodular functions with unknown weights. The problem was first introduced by \cite{yue2011linear}, who studied the problem with the goal of regret minimization. Under this setting, the strategy is to explore new choices of the subsets while also exploiting previous samples to select good ones. Therefore, the algorithm should be encouraged to choose the sets with higher function values while the main focus of our paper is to study the linear submodular bandit under best-arm identification setting, where the objective is to find the best solution set in fewer queries.  
One obvious approach to this problem is to simply take sufficiently many noisy samples so that $f$ is approximated very closely for all sets that are to be queried, and then apply existing algorithms for SM on the approximation of $f$. The downsides of such an approach is that not only is it unnecessary to approximate $f$ so closely when making decisions during an algorithm, but this approach does not exploit the linear structure of $f$. A difficulty with such algorithms is that the large number of noisy samples can make such an approach impractical.

Another related line of work is the problem of identifying the best element in linear bandit (The arms are single elements in the universe.), which is first studied in \cite{soare2014best}. The authors proposed static allocation strategies and adaptive strategies based on successive elimination. Later on, there are a large line of works that proposed more adaptive algorithms \cite{xu2018fully,degenne2020gamification}. \cite{xu2018fully} proposed an algorithm named LinGap which is claimed to be fully adaptive and performed better than those in \cite{soare2014best}. \cite{degenne2020gamification} proposed the first algorithm that is asymptotically optimal.

\section{Preliminaries}
\label{sect:prelim}

In this section, we present the basic setup for our problem, including motivating examples, 
a description of the linear bandit setting, and related technical background that will be used for our algorithms.
First, we provide some additional information about our problem formulation.
We assume that the $l_2$ norm of $\vect{w}$ is bounded by $S$, and the $l_2$ norm of $\vect{F}(s)$ is bounded by $L$ for any $s\in U$. 

 For any $S\subseteq U$ and $x\in U$, the sampling result of the marginal gain $\Delta f(S,x)$ is i.i.d $R$-sub-Gaussian with expectation $\Delta f(S,x)$. Let us denote the $t$-th independent samples of $\Delta f(S,x)$ as $\Delta f_t(S,x)$.  We assume the noisy queries can be expressed as 
$\Delta f_t(S,x)= \sum_{i=1}^dw_i\Delta F_i(S,x) + \xi_t=\Delta\vect{F}^T(S,x)\vect{w}+\xi_t$,
where $\xi_t$ are i.i.d zero-mean noise and $R$-sub-Gaussian. Therefore we have that $\mathbb{E}\Delta f_t(S,x)=\sum_{i=1}^dw_i\Delta F_i(S,x)=\Delta f(S,x)$. Linear submodular functions arise in a wide range of applications including recommender systems, weighted coverage, and facility location. Next, we describe the application of the recommender system as an example to illustrate our problem setting.
\subsection{Motivating application: diversified recommender systems}
 In this application, the user examines
the recommended list of $S$ of $\kappa$ items from the first item to
the last, clicks on the first attractive item, and does not
examine the rest. Herein, the submodular function is the expected total clicks from recommending the list of items. Suppose $F_i(S)$ denotes a submodular function quantifying the incorporation of topic $i$ within the recommended item set $S$ which is usually modeled by the probabilistic coverage model  \cite{yue2011linear,hiranandani2020cascading}, and that $w_i$ is the preference of the user towards the topic $i$ and is unknown to the algorithm. Then $f(S)$ is defined as $f(S)=\sum_{i\in d}w_iF_i(S)$. By definition, we have that $\sum_{i}w_i=1$. Therefore, $||\vect{w}||_2\leq||\vect{w}||_1=1$ and we can set $S=1$. 

 Next, we list necessary notation: (i). Let us denote the marginal gain of adding element $x\in U$ to a set $S\subseteq U$ as $\Delta f(S,x)$, i.e., $\Delta f(S,x):=f(S\cup\{x\})- f(S)$. Therefore, by definition, we have that $\Delta f(S,x)=\sum_{i=1}^dw_i\Delta F_i(S,x)=\Delta \vect{F}(S,x)^T\vect{w}$. Here $\Delta \vect{F}(S,x)=(\Delta F_1(S,x),...,\Delta F_d(S,x))^T$. (ii) When set $S$ is clear from context, we also use $\vect{F}_x$
to denote $\Delta \vect{F}(S,x)$, which is the marginal gain vector of adding
new element $x$ to $S$. Throughout this paper, we define the matrix norm as 
$||\vect{x}||_{\vect{A}}=\sqrt{\vect{x}^T\vect{A}\vect{x}}$. We use $[n]$ to denote the set $\{1,2,...,n\}$. In particular, we use 
$U$ and $[n]$ interchangeably.

\subsection{The linear bandit setting}
\label{sect:lin_bandit}
Finally, we describe the linear bandit problem and how it can be related to our noisy submodular optimization setting. Suppose there are $n$ elements in the universe with features $\vect{x}_1$, $\vect{x}_2$, ...., $\vect{x}_n\in\mathbb{R}^d$. We denote the set of features as $\mathcal{X}=\{\vect{x}_1,\vect{x}_2,...,\vect{x}_n\}$. At each step $t$, the algorithm selects an element $a_t\in [n]$, and observe the reward $r_t=\vect{x}_{a_t}^T\vect{w}+\xi_t$. Here $\vect{w}$ is unknown and we assume $||\vect{w}||_2\leq S$. We also define the arm selection sequence of the first $t$ steps as
 $\vect{X}_t=(\vect{x}_{a_1},\vect{x}_{a_2},...,\vect{x}_{a_t})^T$, and the vector of rewards of the first $t$ steps as $R_t=(r_1,r_2,...,r_t)^T$. A central component of best-arm identification in the linear bandit setting is having an estimate of the unknown vector $\vect{w}$. The regularized least-square estimate of the vector $\vect{w}$ with some regularizer $\lambda$ is thus
\begin{align*}
    \hat{\vect{w}}_t^\lambda=(\sum_{i=1}^t\vect{x}_{a_i}\vect{x}_{a_i}^T+\lambda I)^{-1}(\sum_{i=1}^t\vect{x}_{a_i}r_i) =(\vect{X}_t^T\vect{X}_t+\lambda I)^{-1}(\vect{X}_t^TR_t).
\end{align*} 
The matrix $\vect{X}_t^T\vect{X}_t+\lambda I$ is defined as $\vect{A}_t$ and the vector $\vect{X}_t^TR_t$ is defined as $\vect{b}_t$. Consequently, $\vect{w}_t=\vect{A}_t^{-1}\vect{b}_t$. Notice that here the inverse of $\vect{A}_t$ can be updated quickly by using the Woodbury formula in Lemma \ref{lem:matrix_inversion}.
Here we explain how the linear bandit model can be used in our algorithms. Suppose we choose to sample the marginal gain of adding element $s_t$ to subset $S_t$ at time step $t$, then we can set $\vect{x}_{a_t}=\Delta \vect{F}(S_t,s_t)$, the reward $r_t$ is thus the noisy marginal gain feedback, i.e., $r_t=\Delta f_t(S_t,s_t)=\Delta \vect{F}(S_t,s_t)^T\vect{w}+\xi_t$. 
\subsection{Concentration Properties of Estimation of Weight Vector}
\label{sect:concentration}
Our estimate of $\vect{w}$ is used alongside a concentration inequality that tells us how accurate our estimate is in the form of a confidence interval. In contrast to concentration inequalities such as Hoeffding's which is for a real-valued random variable, the confidence region in the concentration inequality we use can be geometrically interpreted as a ellipsoid. Pulling different arms results in the shrinking of the confidence region along different directions. The following is the concentration inequality that we will use for our least-squares estimator. The result is obtained from Thm. 2 in \cite{abbasi2011improved}. The proof can be found in the appendix. 
\begin{proposition}
\label{prop:adaptive}

    Let $\hat{\vect{w}}_t^{\lambda}$ 
be the solution to the regularized least-squares problem with
regularizer $\lambda$ and let $\vect{A}_t^{\lambda}
 = \vect{X}_t^T\vect{X}_t+\lambda \vect{I}$.  Then for any $N\geq 0$ and every adaptive sequence $\vect{X}_t$ such
that at any step t, $\vect{x}_{a_t}$ only depends on past history, we have that with probability at least $1-\delta$, it holds that
$|\vect{y}^T(\hat{\vect{w}}_t^{\lambda}-\vect{w})|\leq ||\vect{y}||_{(\vect{A}_t^{\lambda})^{-1}}C_t$
for all $t\geq N$ and all $\vect{y}\in \mR^d$ that only depends on past history up to time $N$, where $C_t$ is defined as 
\begin{align*}
C_t=R\sqrt{2\log\frac{\det(\vect{A}_t^{\lambda})^{\frac{1}{2}}\det(\lambda I)^{-\frac{1}{2}}}{\delta}}+\lambda^{\frac{1}{2}}S.
\end{align*}
\end{proposition}

\section{The Standard Greedy Algorithm}
\label{sec:adapt}



In this section, we propose and analyze our first algorithm \alglong (\alg), which is inspired by the standard greedy algorithm with the assumption of access to the exact value oracle to $f$. In such an algorithm, at each round, an element with the highest marginal gain $u=\arg\max_{x\in U}f(S,x)$ is selected and added to the solution set $S$. Since in our problem, $f$ is only accessible via noisy queries, we develop an algorithm inspired by the best-arm-identification algorithm for multi-armed-bandit of \cite{xu2018fully} to select the element to add to the solution set. We provide the algorithm description and overview as follows.

 \textbf{Algorithm overview. }We now describe \alg, pseudocode for which is given in Algorithm \ref{alg:LG}. \alg proceeds in a series of $\kappa$ rounds, with $\kappa$ being the input budget, where during each round an element $u$ of maximum marginal gain is added to the solution with high probability. $S$ is the solution set that \alg builds, and let $S_l$ be the solution set before the $l$-th element is added at the end of round $l$. The element to be added is determined by making a series of noisy queries, i.e. taking samples of noisy marginal gains.
 We can view the problem of finding $a\in U$ such that $\Delta f(S,a)\geq \max_{u\in U}\Delta f(S,u)$ as a linear bandit problem with $n$ arms corresponding to the elements of $U$. The set of arms from which we take samples includes all past marginal gains. Our objective is to find an arm $\hat{a}$ that satisfies $\vect{x}^T_{\hat{a}}\vect{w}\geq\max_{u\in [n]}\vect{x}^T_{u}\vect{w}-\epsilon'$ where $\epsilon'=\epsilon/\kappa$. \alg keeps track of the following throughout its duration: (i) A variable $t$ indicating the total number of samples that have been taken since the beginning of the algorithm, (ii) the matrix $\vect{A}$ and the vector $\vect{b}$ for the least-squares estimate of $\vect{w}$ as described in Section \ref{sect:lin_bandit}, and (iii) the total number of noisy samples of the marginal gain $\Delta f(S_{l'},j)$ for each $l'\in\{1,2,3...,l\}$ and $j\in [n]$ denoted as $T_{l',u}$.

We now describe a round of \alg in detail. 
The round begins by taking a single noisy sample of the marginal gain of adding each element to the current solution $S$ (Lines \ref{line:LBAS_for_loop_starts} to \ref{line:LBAS_for_loop_ends}). Next, \alg proceeds to a loop (Lines \ref{line:LBAS_while_loop_starts} to \ref{line:LBAS_while_loop_ends}) where each iteration corresponds to a single noisy marginal gain sample, and the loop exits once the highest marginal gain element is identified with high probability (when Line \ref{line:check_stopping_condition} is satisfied). At the beginning of the loop, the guess of $\vect{w}$ is computed based on the procedure described in Section \ref{sect:lin_bandit} (Line \ref{line:compute_w_in_LG}). Next in the loop, \alg identifies the best marginal gain vector to sample from the current and past set of marginal gains, i.e. the set $\{\Delta \vect{F}(S_{l'},a)\}_{l'\in[l],a\in[n]}$ (Lines \ref{line:LBAS_choose_sample_starts} to \ref{line:LBAS_choose_sample_ends}). This is done as follows: Two elements $i_t$ and $j_t$ from the universe $U$ are selected where $i_t$ is the element with the highest estimated marginal gain according to our current estimate of $\vect{w}$ (Line \ref{line:LBAS_choose_sample_starts}), and $j_t$ is the element whose upper confidence bound of the direction $\Delta\vect{F}(S,j)-\Delta\vect{F}(S,i_t)$ is the highest. The upper confidence bound is given by $\Delta\vect{F}(S,j_t)^T\hat{\vect{w}}_t-\Delta\vect{F}(S,i_t)^T\hat{\vect{w}}_t+\beta_t(j_t,i_t)$ where $\beta_t(i,j)=C_t||\Delta\vect{F}(S,j)-\Delta\vect{F}(S,i)||_{\vect{A}_t^{-1}}$ and $C_t=(R\sqrt{2\log\frac{\det(\vect{A}_t^{\lambda})^{\frac{1}{2}}\det(\lambda I)^{-\frac{1}{2}}}{\delta}}+\lambda^{\frac{1}{2}}S)$. 
We then follow a strategy proposed by \cite{xu2018fully}, and discussed in the appendix in Section \ref{appdx:discuss_p^*_greedy}, in order to select the marginal gain vector that makes the confidence interval $||\Delta \vect{F}(S,i)-\Delta \vect{F}(S,j)||_{\vect{A}_t^{-1}}$ to shrink as fast as possible. In particular, we choose the asymptotic-optimal ratio for decreasing the confidence interval $||\Delta \vect{F}(S,i)-\Delta \vect{F}(S,j)||_{\vect{A}_t^{-1}}$ when the number of samples goes to infinity. We denote the sampling ratio of the marginal gain vector $\Delta \vect{F}(S_{l'},a)$ at the round $l$ for estimating the direction of $\Delta \vect{F}(S,i)-\Delta \vect{F}(S,j)$ as $p^*(l',a|l,i,j)$. We pull the arm so that the number of samples of arms is close to the ratio $p$. More specifically, 
\begin{align*}
(l_t,a_t)=\arg\min_{l'\in[l],a\in[n]:p^*_{l',a}(i_t,j_t)>0}\frac{T_{l',a}}{p^*(l',a|l,i_t,j_t)}.
\end{align*}

This concludes the description of our algorithm \alg.

 


\textbf{Technical challenges. } We now discuss the technical difficulties of developing \alg. It should be noted that the portion of \alg where we are repeatedly making noisy queries in order to determine the element of approximately highest marginal gain with high probability could be replaced by any existing algorithm for $\epsilon$-approximate best-arm-identification in linear bandit where the arms correspond to the marginal gains of adding each of the $n$ elements of $U$ to the current solution $S$. In this case, each round of \alg would be a new instance of linear bandit. In particular, we develop an algorithm using the static allocation strategy as the subroutine in Section \ref{appdx:static}. However, such algorithms overlook that the weight vector estimated in different rounds of the algorithm are the same and don't reuse samples gained from the past rounds while our algorithm \alg is different because we do not view each round of \alg as a new instance of linear bandit, but instead we use information gained about $\vect{w}$ from past rounds of the greedy algorithm. In addition, \alg may sample marginal gains corresponding to arms of the previous rounds of the greedy algorithm. Thus, our contribution extends beyond merely integrating best-arm identification into the greedy algorithm. We also develop an algorithm, inspired by \cite{xu2018fully}, which efficiently reuses samples from previous rounds, thereby enhancing the algorithm’s performance and efficiency.


\textbf{Theoretical Guarantees. }
Finally, we present the theoretical guarantees of \alg. Here we denote the total number of samples of $\Delta \vect{F}(S_{l'},a)$ before iteration $l$ as $N_{l',a}^{(l)}$, and the total number of samples before iteration $l$ as $N^{(l)}=\sum_{l'\in[l],a\in[n]}N_{l',a}^{(l)}$. Moreover, we define $\Delta_{l,i}=\max_{a\in[n]}\Delta \vect{F}(S_{l},a)^T\vect{w}-\Delta\vect{F}(S_{l},i)^T\vect{w}$. Then we have the following results. The proof is deferred to Section \ref{appdx:adapt} in the appendix.

\begin{algorithm}[t]
\caption{\alglong (\alg)}\label{alg:LG}
 \begin{algorithmic}[1]
 \STATE \textbf{Input:} $\delta, \epsilon$
 \STATE \textbf{Output: } solution set $S\subseteq U$
 \STATE  $S\gets \emptyset$
 \STATE  $\vect{A}\gets\lambda \vect{I}$, $\vect{b}\gets\vect{0}$
 \STATE $t\gets1$
 \FOR{$l=1,...,\kappa$}
\FOR{$a\in[n]$}\label{line:LBAS_for_loop_starts}
\STATE sample $r_t=\Delta \vect{F}^T(S,a)\vect{w}+\xi_t$ and obtain reward $r_t$
\STATE Update $\vect{A}\gets \vect{A}+\Delta \vect{F}(S,a)\Delta \vect{F}^T(S,a)$ and $\vect{b}\gets \vect{b}+r_t\Delta \vect{F}(S,a)$\label{line:LBAS_update_A_b1}
\STATE $t\gets t+1$
\STATE $T_{l,a}\gets1$
\ENDFOR \label{line:LBAS_for_loop_ends}
  \STATE $S_{l}\gets S$
  \WHILE{\textbf{true}}\label{line:LBAS_while_loop_starts}
  \STATE $\hat{\vect{w}}_t=\vect{A}^{-1}\vect{b}$\label{line:compute_w_in_LG}
 \STATE $i_t\gets\arg\max_{i\in[n]}\Delta\vect{F}(S,i)^T\hat{\vect{w}}_t$\label{line:LBAS_choose_sample_starts}
 \STATE $j_t\gets\arg\max_{j\in[n]}\Delta\vect{F}(S,j)^T\hat{\vect{w}}_t-\Delta\vect{F}(S,i_t)^T\hat{\vect{w}}_t+\beta_t(j,i_t)$
 \STATE $B(t)\gets\Delta\vect{F}(S,j_t)^T\hat{\vect{w}}_t-\Delta\vect{F}(S,i_t)^T\hat{\vect{w}}_t+\beta_t(j_t,i_t)$
 \STATE $(l_t,a_t)=\arg\min_{l'\in[l],a\in[n]:p^*_{l',a}(i_t,j_t)>0}\frac{T_{l',a}}{p^*(l',a|l,i_t,j_t)}$\label{line:LBAS_choose_sample_ends}
 \IF{ $B(t)\leq\epsilon/\kappa$}\label{line:check_stopping_condition}
  \STATE \textbf{return} $i_t$
 \ENDIF
\STATE sample $r_t=\Delta\vect{F}^T(S_{l_t},a_t)\vect{w}+\xi_t$
 \STATE Update $\vect{A}\gets \vect{A}+\Delta \vect{F}(S_{l_t},a_t)\Delta \vect{F}^T(S_{l_t},a_t)$ and $\vect{b}\gets \vect{b}+r_t\Delta \vect{F}(S_{l_t},a_t)$
 \STATE $T_{l_t,a_t}\gets T_{l_t,a_t}+1$
  \STATE $t\gets t+1$
 \ENDWHILE\label{line:LBAS_while_loop_ends}

 \STATE $S\gets S\cup \{u\}$

 \ENDFOR
 \STATE \textbf{return} $S$
 \end{algorithmic}
\end{algorithm}


\begin{theorem}
\label{thm:greedy}
    With probability at least $1-\delta$, the following statements hold: 
     \begin{enumerate}
        \item The exact function value of the output solution set $S$ satisfies that $f(S)\geq(1-e^{-1})f(OPT)-\epsilon.$ Here $OPT$ is an optimal solution to the SM problem;
         
        \item During each round in \alg, as the number of samples increases, the confidence region would shrink, and the stopping condition $B(t)\leq \epsilon/\kappa$ would be met ultimately. Assuming $\lambda\leq\frac{2R^2}{S^2}\log\frac{1}{\delta}$, 
        then \alg at round $l$ takes at most
    \begin{align*}
    \sum_{l'\in[l],a\in[n]}\max\{4H^{(l)}_{l',a} R^2(2\log\frac{1}{\delta}+d\log M^{(l)})+1-N_{l',a}^{(l)},0\}
\end{align*}
    number of samples, where $M^{(l)}=\frac{16H_\epsilon^{(l)} R^2L^2}{\lambda}\log(\frac{1}{\delta})+\frac{32(H_\epsilon^{(l)})^2 R^4L^4}{\lambda^2 }+\frac{2(N^{(l)}+nl)L^2}{\lambda d}+2$. Here $H_{l',a}^{(l)}=\max_{i,j\in[n]}\frac{p^*(l',a|l,i,j)\rho_{i,j}^l}{\max\{\epsilon/\kappa,\frac{\Delta_{l,i}\vee\Delta_{l,j}+\epsilon/\kappa}{3}\}^2}$ and $H_{\epsilon}^{(l)}=\sum_{l'\in[l],a\in[n]}H_{l',a}^{(l)}$, where $\rho_{i,j}^l$ is the optimal value of (\ref{eqn:optimization}).
    \end{enumerate}
\end{theorem}
Now we discuss the sample complexity. First of all, if we set $N_{l',a}^{(l)}=0$, then the result is reduced to the case where we don't utilize the samples of the marginal gains from the past rounds. Compared with this case, our sample complexity of using past data can be improved by at most $N_{l',a}^{(l)}$ from the past rounds for each $l',a$ while only sacrificing the factor of $O(\log(N^{(l)}))$. On the other hand, as discussed in the algorithm overview, selecting an element to add to the solution set corresponds to a best-arm identification problem in linear bandits. Thus, we can compare our results with the element-selecting procedure being replaced by LinGapE in \cite{xu2018fully}, which does not leverage samples of noisy marginal gains from previous rounds of the greedy algorithm.  
 The sample complexity of  LinGapE is  $\tau_2=4H_\epsilon^{(l)} R^2(2\log\frac{\kappa n^2}{\delta}+d\log M^{(l)})+n$, which is worse than the sample complexity of \alg in the case where we don't utilize samples from the past, and the set of samples we can take is limited to the set of $\{\Delta \vect{F}(S,a)\}_{a\in[n]}$ by a factor of $O(\log n^2)$.

 On the other hand, we have to notice that to find the asymptotic sample selection ratio $p^*(l',a|l,i,j)$, the algorithm involves solving a linear programming problem for each $i,j$ at each round $l$ to find $\{p^*(l',a|l,i,j)\}_{l'\in[l],a\in[n]}$, which results in solving a total of $O(\kappa n^2)$ number of the optimization problems. In particular, if we replace the allocation ratio $p^*$ with any other ratio $p$, the result in the theorem would still holds, with the value of $\rho^{l}_{i,j}$ being replaced by $||\Delta\vect{F}(S,i)-\Delta\vect{F}(S,j)||^2_{\Lambda_{i,j}^{-1}}$ where ${\Lambda_{i,j}^{-1}}$ is defined in (\ref{eqn:define_lambda}).

\section{Threshold Greedy with Adaptive Allocation strategy}
\label{sec:threshold}
In the last section, we present our algorithm \alg which is based on the standard greedy algorithm but with linear bandit feedback. However, the implementation of \alg poses several challenges in terms of its scalability: (i) Finding the allocation strategy in \alg necessitates solving $O(\kappa n^2)$ linear programming problems, (ii) \alg requires updating the inverse of matrix $\vect{A}$ and confidence interval every time an arm is pulled, (iii) \alg is sensitive to the gap between the marginal gains of different elements and therefore may need many samples in the case that there exist elements that have an expected reward close to optimal, (iv) the standard greedy algorithm is less efficient compared to alternative greedy algorithms even in the exact value oracle setting. Motivated by these weaknesses of \alg, we introduce an alternative algorithm \algthrelong (\algthre) which is based on the threshold greedy algorithm of \cite{badanidiyuru2014fast}, but incorporates a linear bandit strategy.

We now describe \algthre, pseudocode for which is given in Algorithm \ref{alg:ATG}. \algthre proceeds in a series of rounds (Lines \ref{line:tg_for_loop_starts} to \ref{line:tg_for_loop_ends}), each round corresponding to a threshold value $w$.
$w$ is initialized to be an approximation of the maximum $f$ value of a singleton in $U$.
During each round of \algthre, we iterate over the ground set $U$ and make a decision about whether to add each element $a\in U$ to our solution $S$ or not depending on whether the marginal gain $\Delta f(S,a)$ is above $w$. \algthre decides whether to add each $a\in U$ to $S$ or not on Lines \ref{line:check_cardi} to \ref{line:end_eval} based on a sampling procedure in a linear bandit-like setting, and is analogous to the best-arm-identification procedure in \alg. Throughout the algorithm, we keep track of the following parameters: the covariance matrix $\vect{A}$, vector $\vect{b}$, and the set of marginal gain vectors to sample from, which is denoted as $\mathcal{X}$. Here we denote the size of $\mathcal{X}$ as $m$. 
In this algorithm, $\mathcal{X}$ includes all the past marginal gains that have been evaluated. Let us denote $\mathcal{X}=\{\Delta\vect{F}(S_i,a_i)\}_{i\in[m]}$ where $S_i$ and $a_i$ are the $i$-th evaluated solution set and element.
\algthre first begins by taking a single sample of the marginal gain of the current element of $U$ (Line \ref{line:tg_init_sample_starts}). 
In a loop on Lines \ref{line:tg_while_loop_starts} to \ref{line:tg_while_loop_ends}, \algthre iteratively identifies the best marginal gain vector to sample from the set $\mathcal{X}$ in order to decrease $||\Delta\vect{F}(S,a)||_{\vect{A}^{-1}}$ as fast as possible. Notice that this is different from the strategy of \alg since we are no longer attempting to distinguish the highest marginal gain from the rest, but only make a decision about a single marginal gain. Let $p_i^*$ denote the fraction of samples allocated to element $\Delta\vect{F}(S_i,a_i)$, with $\sum_{i\in[m]}p_i=1$, which is asymptotically optimal for decreasing $||\Delta\vect{F}(S,a)||_{\vect{A}^{-1}}$. $p_i^*$ can be calculated by solving a linear programming problem (see Appendix for more details about how to calculate $p^*_i$). Notice that there are a total of $O(\frac{n}{\alpha}\log\frac{\kappa}{\alpha})$ number of evaluations of the marginal gains, and for each marginal gain, we only need to solve one linear programming problem. Thus \algthre only requires solving $O(\frac{n}{\alpha}\log\frac{\kappa}{\alpha})$ linear programming problems which decreases compared with \alg. On the other hand, since we only consider one allocation ratio for each evaluated marginal gain, the greedy selection strategy in Line \ref{line:find_sample} can be changed to run in batches. Here $\beta_t$ denotes the confidence interval which is defined as $\beta_t=\big(R\sqrt{2\log\frac{2\det(\vect{A}_t)^{\frac{1}{2}}\det(\lambda I)^{-\frac{1}{2}}}{\delta}}+\lambda^{\frac{1}{2}}S\big)||\Delta \vect{F}(S,a)||_{\vect{A}_t^{-1}}$. The theoretical guarantee of \algthre is in Theorem \ref{thm:threshold}.

  
Each iterative sample is followed by an update of $\vect{A},\vect b$, and then the estimate $\hat{\vect w}$ are all updated according to the procedure described in Section \ref{sect:lin_bandit} on Lines \ref{line:updateAb_threshold}. Once it has been determined whether the marginal gain is above $w$ (Line \ref{line: comparison to thres 1}) or not (Line \ref{line: comparison to thres 2}), $a$ is added or not respectively to $S$ and \algthre repeats the procedure with the next element of $U$.

 \begin{algorithm}[t]
\caption{\algthrelong (\algthre)}\label{alg:ATG}
 \begin{algorithmic}[1]
 \STATE \textbf{Input:} $\epsilon$, $\delta, \alpha$
 \STATE $\mathcal{X}\gets\emptyset$,  $N_0\gets\frac{2R^2}{\epsilon^2}\log(6n/\delta)$, 
 \STATE $\vect{A}\gets\lambda \vect{I}$, $\vect{b}\gets\vect{0}$
 \FORALL{$a\in [n]$}
 \STATE $\hat{f}(a) \gets $ sample average over $N_0$ samples with features $ \vect{F}(a)$ \label{alg:ATG:line:sample-mean}
 \STATE $\vect{A}\gets \vect{A}+N_0\vect{F}(a)\vect{F}^T(a)$, $\vect{b}\gets \vect{b}+ N_0\hat{f}(a)\vect{F}(a)$
 \STATE $T_{\emptyset, a}=N_0$, $\mathcal{X}\gets \mathcal{X}\cup\{\vect{F}(a)\}$
 \ENDFOR
  
  \STATE $g:=\max_{a\in [n]}\hat{f}(a)$, 
 \STATE  $S\gets \emptyset$
 \STATE $t\gets nN_0$, $m=n$
 \FOR{$w=g$; $w=(1-\alpha)w$; $w>\alpha g/\kappa$}\label{line:tg_for_loop_starts}
 \FORALL{$a\in [n]$} 
\IF{$|S|<\kappa$}\label{line:check_cardi}
\STATE sample $r_t=\Delta\vect{F}^T(S,a)\vect{w}+\xi_t$ and obtain the reward $r_t$\label{line:tg_init_sample_starts}
\STATE update $\vect{A}$ and $\vect{b}$, $t\gets t+1$ 
 \STATE $\hat{\vect{w}}=\vect{A}^{-1}\vect{b}$
    \STATE$\mathcal{X}\gets\mathcal{X}\cup\{\Delta \vect{F}(S,a)\}$, $m=m+1$
    \STATE $S_m\gets S$, $a_m\gets a$
    \STATE $T_{S_m,a_m}\gets1$
    \WHILE{\textbf{true}}\label{line:tg_while_loop_starts}
 \STATE $i_t=\arg\min_{i\in[|\mathcal{X}|]:p_i^*>0}\frac{T_{S_i,a_i}}{p_i^*}$\label{line:find_sample}
        \IF{$\Delta \vect{F}(S,a)^T\hat{\vect{w}}-\beta_t \geq w-\epsilon$} 
        \label{line: comparison to thres 1}
            \STATE $S\gets S\cup \{a\}$; \textbf{break}
        \ELSIF{$\Delta \vect{F}(S,a)^T\hat{\vect{w}}+\beta_t \leq w+\epsilon$} 
        \label{line: comparison to thres 2}
            \STATE \textbf{break}
        \ENDIF
        \STATE sample $r_t=\Delta\vect{F}^T(S_{i_t},a_{i_t}))\vect{w}+\xi_t$
 \STATE update $\vect{A}$ and $\vect{b}$, $\hat{\vect{w}}=\vect{A}^{-1}\vect{b}$\label{line:updateAb_threshold}
 \STATE $T_{S_{i_t},a_{i_t}}\gets T_{S_{i_t},a_{i_t}}+1$
  \STATE $t\gets t+1$
 \ENDWHILE\label{line:tg_while_loop_ends}
  \ENDIF\label{line:end_eval}
 \ENDFOR

 \ENDFOR\label{line:tg_for_loop_ends}
 \STATE \textbf{return} $S$
 \end{algorithmic}
\end{algorithm}

    

\begin{theorem}
\label{thm:threshold}
     \algthre makes $n\log(\kappa/\alpha)/\alpha$ evaluations of the marginal gains. In addition, with probability at least $1-\delta$, the following statements hold: 
     \begin{enumerate}
        \item The function value of the output solution set $S$ satisfies that $f(S)\geq(1-e^{-1}-\alpha)f(OPT)-2\epsilon.$ Here $OPT$ is an optimal solution to the SM problem;
         
        \item  Assuming $\lambda\leq\frac{2R^2}{S^2}\log\frac{2}{\delta}$, the $m$-th evaluation of the marginal gain of adding an element $a$ to $S$ in \algthre takes at most
    \begin{align*}
        \sum_{i\in[m]}\max\{4H^{(m)}_{i} R^2(2\log\frac{2}{\delta}+d\log M^{(m)})+1-N_{i}^{(m)},0\}
\end{align*}
  samples, where $M^{(m)}=\frac{16H_\epsilon^{(m)} R^2L^2}{\lambda}\log\frac{2}{\delta}+\frac{32(H_\epsilon^{(m)})^2 R^4L^4}{\lambda^2 }+\frac{2(N^{(m)}+m)L^2}{\lambda d}+2$. 
  Here $H_{i}^{(m)}=\frac{p^*_{i}\rho^{(m)}}
  {\max\{\frac{\epsilon/\kappa+|w-\Delta \vect{F}(S,a)^T\vect{w}|}{2},\epsilon/\kappa\}^2}$ 
  and $H_{\epsilon}^{(m)}=\sum_{i\in[m]}H_{i}^{(m)}$. 
  $p^*_{i}$ is the fraction of samples allocated to sample $\Delta \vect{F}(S_i,a_i)$ and therefore $\sum_{i\in[m]}p^*_{i}=1$. $N^{(m)}$ is the total number of samples before the $m$-th evaluation of marginal gains. $N^{(m)}_i$ is the number of samples to the marginal gain $\Delta f(S_i,a_i)$ before the $m$-th evaluation of marginal gains. $\rho^{(m)}$ is the optimal vale of (\ref{eqn:opt_tg}).
    \end{enumerate}
\end{theorem}
The result on the sample complexity can be derived directly by Theorem \ref{thm:samp}. The detailed proof of the approximation ratio is presented in Section \ref{appdx:threshold} in the appendix. If we replace the allocation ratio $p^*$ with any other ratio $p$, the result in the theorem would still hold, with the value of $\rho^{(m)}$ being replaced by $||\Delta\vect{F}(S_m,a_m)||^2_{\Lambda^{-1}}$ where $\Lambda=\sum_{i\in[m]}p_i^*\Delta\vect{F}(S_i,a_i)\Delta\vect{F}(S_i,a_i)^T$.
\section{Experiments}
\label{sec:exp}
 
In this section, we conduct experiments on the instance of movie recommendation. The dataset used in the experiments are subsets extracted from the MovieLens 25M dataset \cite{harper2015movielens}, which comprises $162,541$ ratings for $13,816$ movies, each associated with $1,128$ topics. We compare the sample complexity of different algorithms with different values of size constraint $\kappa$, error parameter $\epsilon$, and different datasets with different values of $d$. When $\kappa$ is varied, $\epsilon$ is fixed at $0.1$. The dataset movie60 contains $60$ elements with $d=5$, $|V|=500$. The dataset movie5000 contains $n=5000$ elements with $d=30$, $|V|=1000$. For the algorithms comparing different values of $d$, we run the algorithms on different datasets with different values of $d$. Here $n$ and $|V|$ are fixed at $500$. Additional details about the applications, setup, and results can be found in Section \ref{appdx:exp} in the supplementary material.

In the application of movie recommendation, the submodular basis function $F_i$ denotes how well a subset movies the topic $i$, which is defined by the probabilistic movieage model as is presented in the appendix. The noisy marginal gain is sampled in the following way: first we uniformly sample a user $a_t$ from the set of users $V$, then the noisy marginal gain of adding a new movie $x$ into a subset $S$ is $\Delta f_t(S,x)=\sum_{i\in[d]}w(a_t,i)\Delta F_i(S,x)=\vect{w}(a_t,\cdot)^T\Delta \vect{F}(S,x)$. Consequently, the exact value of our submodular objective is the average of marginal gain for all elements, i.e., 
$\Delta f(S,x)= \frac{1}{|V|}\sum_{a\in V}\vect{w}(a,\cdot)^T\Delta \vect{F}(S,x)$. Denote the average weight vector $\bar{\vect{w}}$ as $\bar{\vect{w}}=\frac{1}{|V|}\sum_{a\in V}\vect{w}(a,\cdot)$, then it follows that
$\Delta f(S,x)= \bar{\vect{w}}^T\Delta \vect{F}(S,x)$. 

We compare the solutions returned by the following five algorithms: (i). The algorithm \alg ("Lin-GREEDY"). Due to the high computational cost, here we don't solve the linear programming problems and replace the asymptotic optimal allocation ratio $p^*$ with the allocation $p$ that satisfies $p(l',a|l,i,j)=0.5$ if $a$ is $i$ or $j$ and $l'=l$. Otherwise $p(l',a|l,i,j)=0$ (see Section \ref{appdx:discuss_p^*_greedy} for more details.) ; (ii). The linear greedy algorithm \algthre. Here we consider two choices of the allocation parameters. The first one satisfies that $p_m=1$ and $p_a=0$ for $a\neq m$, i.e., the algorithm only samples the current marginal gain ("LinTG-H"). The second one is $p_a=\frac{|w_a^*|}{\sum_{a\in[m]}|w_a^*|} $ where $w_a^*$ is the optimal solution of the linear programming problem defined in the algorithm description ("LinTG");  (iii). The ExpGreedy algorithm from \cite{singla2016noisy} with the parameter $k'$ set to be $1$ ("Exp-GREEDY"); (iv). The threshold-greedy algorithm without considering the linear structure ("TG"). More details about the algorithms can be found in the appendix.

\begin{figure*}[t!]
    \centering
    \hspace{-0.5em}
     \subfigure[movie60 $\kappa$]
{\label{fig:movie60k}\includegraphics[width=0.24\textwidth]{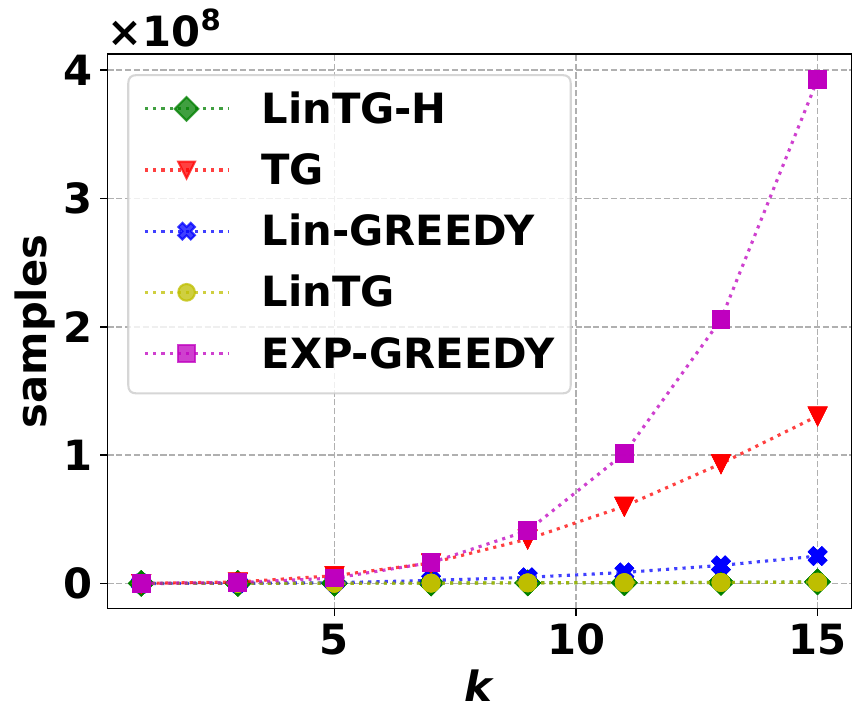}} 
\hspace{-0.5em}
     \subfigure[movie5000 $\kappa$]
{\label{fig:movie5000k}\includegraphics[width=0.24\textwidth]{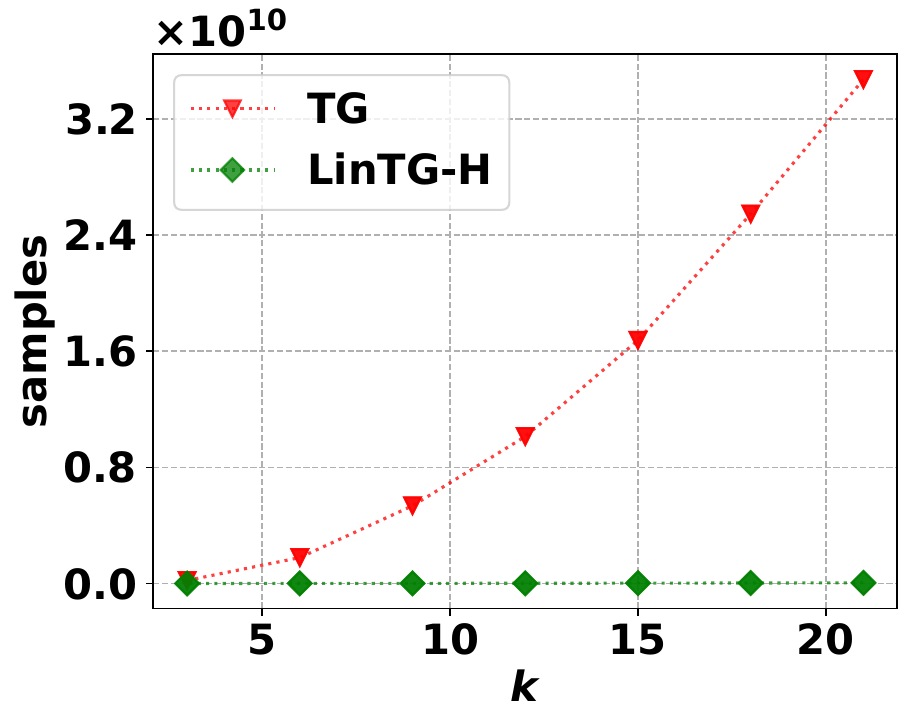}} 
    \hspace{-0.5em}
     \subfigure[movie60 $\epsilon$]
{\label{fig:movie60eps}\includegraphics[width=0.24\textwidth]{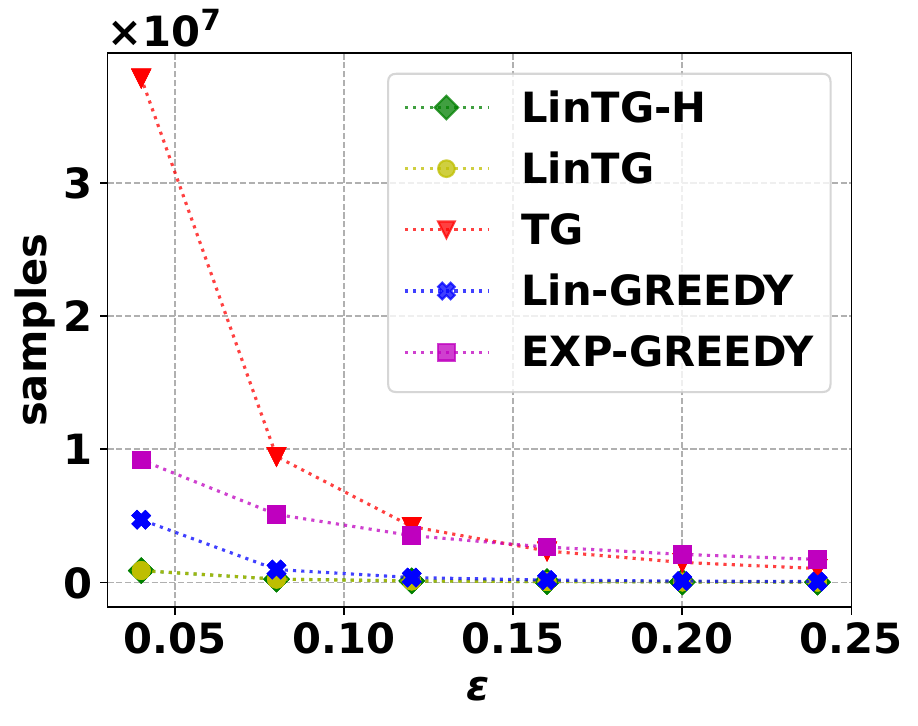}} 
\hspace{-0.5em}
\subfigure[movie500 d]
{\label{fig:movie500d}\includegraphics[width=0.24\textwidth]{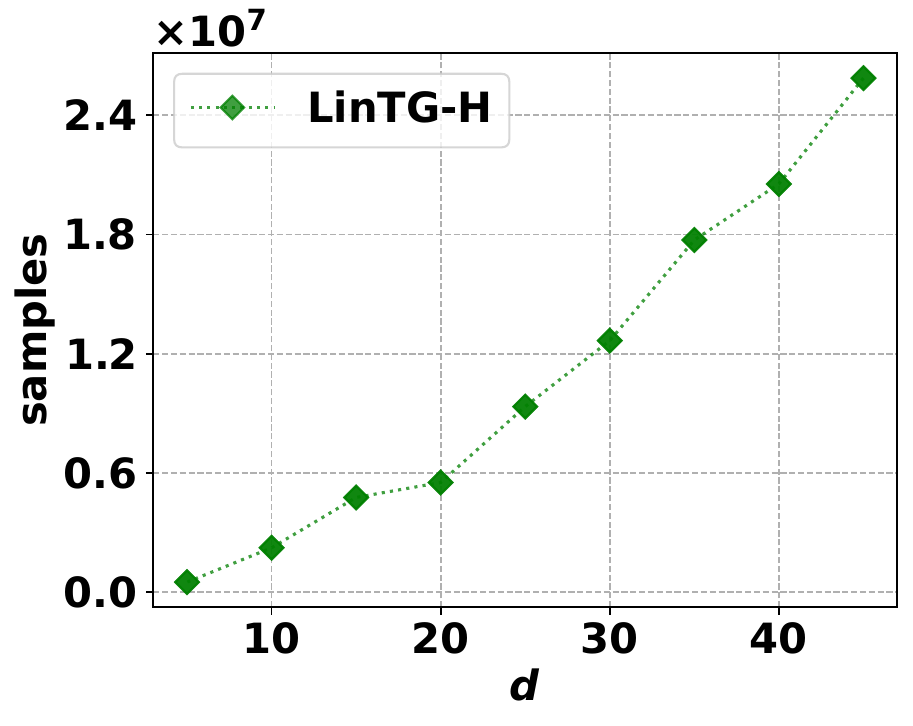}}
\hspace{-0.5em}
\caption{The experimental results of running the algorithms on instances of movie recommendation on the subsets of MovieLens 25M dataset with $n=60$, $d=5$ ("movie60") and $n=5000$, $d=30$ ("movie5000"), and different datasets with different values of $d$.}
        \label{fig}
\end{figure*}

We present the results in terms of the number of evaluations in Figure \ref{fig}. Due to the high sample complexity and heavy computational burden resulting from solving the linear programming problem, here we only run the algorithms EXP-GREEDY, LinGREEDY, and LinTG on the relatively small dataset movie60. From the results on the dataset movie60, we can see that the sample complexity of the algorithms that don't consider the linear structure (EXP-GREEDY and TG) is much higher than the algorithms proposed in this paper, which demonstrates the advantages of our proposed methods. In particular, the algorithms LinTG and LinTG-H have the lowest sample complexity compared with other methods. Besides, we notice that the algorithm when $\epsilon$ is very small, the algorithms TG has an increase on the sample complexity. From Figure \ref{fig:movie500d}, we can see the sample complexity of the algorithm LinTG-H increases as $d$ increases, which is consistent with our theoretical results on sample complexity.

\bibliographystyle{plainnat}
\bibliography{ref}

\begin{thebibliography}{30}
\providecommand{\natexlab}[1]{#1}
\providecommand{\url}[1]{\texttt{#1}}
\expandafter\ifx\csname urlstyle\endcsname\relax
  \providecommand{\doi}[1]{doi: #1}\else
  \providecommand{\doi}{doi: \begingroup \urlstyle{rm}\Url}\fi

\bibitem[Abbasi-Yadkori et~al.(2011)Abbasi-Yadkori, P{\'a}l, and
  Szepesv{\'a}ri]{abbasi2011improved}
Yasin Abbasi-Yadkori, D{\'a}vid P{\'a}l, and Csaba Szepesv{\'a}ri.
\newblock Improved algorithms for linear stochastic bandits.
\newblock \emph{Advances in neural information processing systems}, 24, 2011.

\bibitem[Badanidiyuru and Vondr{\'a}k(2014)]{badanidiyuru2014fast}
Ashwinkumar Badanidiyuru and Jan Vondr{\'a}k.
\newblock Fast algorithms for maximizing submodular functions.
\newblock In \emph{Proceedings of the twenty-fifth annual ACM-SIAM symposium on
  Discrete algorithms}, pages 1497--1514. SIAM, 2014.

\bibitem[Bateni et~al.(2017)Bateni, Esfandiari, and Mirrokni]{bateni2017almost}
MohammadHossein Bateni, Hossein Esfandiari, and Vahab Mirrokni.
\newblock Almost optimal streaming algorithms for coverage problems.
\newblock In \emph{Proceedings of the 29th ACM Symposium on Parallelism in
  Algorithms and Architectures}, pages 13--23, 2017.

\bibitem[Chen et~al.(2017)Chen, Krause, and Karbasi]{chen2017interactive}
Lin Chen, Andreas Krause, and Amin Karbasi.
\newblock Interactive submodular bandit.
\newblock \emph{Advances in Neural Information Processing Systems}, 30, 2017.

\bibitem[Chen et~al.(2014)Chen, Lin, King, Lyu, and
  Chen]{chen2014combinatorial}
Shouyuan Chen, Tian Lin, Irwin King, Michael~R Lyu, and Wei Chen.
\newblock Combinatorial pure exploration of multi-armed bandits.
\newblock \emph{Advances in neural information processing systems}, 27, 2014.

\bibitem[Crawford et~al.(2019)Crawford, Kuhnle, and
  Thai]{crawford2019submodular}
Victoria Crawford, Alan Kuhnle, and My~Thai.
\newblock Submodular cost submodular cover with an approximate oracle.
\newblock In \emph{International Conference on Machine Learning}, pages
  1426--1435. PMLR, 2019.

\bibitem[Degenne et~al.(2020)Degenne, M{\'e}nard, Shang, and
  Valko]{degenne2020gamification}
R{\'e}my Degenne, Pierre M{\'e}nard, Xuedong Shang, and Michal Valko.
\newblock Gamification of pure exploration for linear bandits.
\newblock In \emph{International Conference on Machine Learning}, pages
  2432--2442. PMLR, 2020.

\bibitem[El-Arini and Guestrin(2011)]{ElAriniG11}
Khalid El-Arini and Carlos Guestrin.
\newblock Beyond keyword search: discovering relevant scientific literature.
\newblock In \emph{Proceedings of the 17th ACM SIGKDD international conference
  on Knowledge discovery and data mining}, pages 439--447. ACM, 2011.

\bibitem[Even-Dar et~al.(2002)Even-Dar, Mannor, and Mansour]{even2002pac}
Eyal Even-Dar, Shie Mannor, and Yishay Mansour.
\newblock Pac bounds for multi-armed bandit and markov decision processes.
\newblock In \emph{Computational Learning Theory: 15th Annual Conference on
  Computational Learning Theory, COLT 2002 Sydney, Australia, July 8--10, 2002
  Proceedings 15}, pages 255--270. Springer, 2002.

\bibitem[Harper and Konstan(2015)]{harper2015movielens}
F~Maxwell Harper and Joseph~A Konstan.
\newblock The movielens datasets: History and context.
\newblock \emph{Acm transactions on interactive intelligent systems (tiis)},
  5\penalty0 (4):\penalty0 1--19, 2015.

\bibitem[Hassidim and Singer(2017{\natexlab{a}})]{hassidim2017robust}
Avinatan Hassidim and Yaron Singer.
\newblock Robust guarantees of stochastic greedy algorithms.
\newblock In \emph{International Conference on Machine Learning}, pages
  1424--1432. PMLR, 2017{\natexlab{a}}.

\bibitem[Hassidim and Singer(2017{\natexlab{b}})]{hassidim2017submodular}
Avinatan Hassidim and Yaron Singer.
\newblock Submodular optimization under noise.
\newblock In \emph{Conference on Learning Theory}, pages 1069--1122. PMLR,
  2017{\natexlab{b}}.

\bibitem[Hiranandani et~al.(2020)Hiranandani, Singh, Gupta, Burhanuddin, Wen,
  and Kveton]{hiranandani2020cascading}
Gaurush Hiranandani, Harvineet Singh, Prakhar Gupta, Iftikhar~Ahamath
  Burhanuddin, Zheng Wen, and Branislav Kveton.
\newblock Cascading linear submodular bandits: Accounting for position bias and
  diversity in online learning to rank.
\newblock In \emph{Uncertainty in Artificial Intelligence}, pages 722--732.
  PMLR, 2020.

\bibitem[Horel and Singer(2016)]{horel2016maximization}
Thibaut Horel and Yaron Singer.
\newblock Maximization of approximately submodular functions.
\newblock \emph{Advances in neural information processing systems}, 29, 2016.

\bibitem[Kalyanakrishnan et~al.(2012)Kalyanakrishnan, Tewari, Auer, and
  Stone]{kalyanakrishnan2012pac}
Shivaram Kalyanakrishnan, Ambuj Tewari, Peter Auer, and Peter Stone.
\newblock Pac subset selection in stochastic multi-armed bandits.
\newblock In \emph{ICML}, volume~12, pages 655--662, 2012.

\bibitem[Karimi et~al.(2017)Karimi, Lucic, Hassani, and
  Krause]{karimi2017stochastic}
Mohammad Karimi, Mario Lucic, Hamed Hassani, and Andreas Krause.
\newblock Stochastic submodular maximization: The case of coverage functions.
\newblock \emph{Advances in Neural Information Processing Systems}, 30, 2017.

\bibitem[Kempe et~al.(2003)Kempe, Kleinberg, and Tardos]{kempe2003maximizing}
David Kempe, Jon Kleinberg, and {\'E}va Tardos.
\newblock Maximizing the spread of influence through a social network.
\newblock In \emph{Proceedings of the ninth ACM SIGKDD international conference
  on Knowledge discovery and data mining}, pages 137--146, 2003.

\bibitem[Leskovec et~al.(2007)Leskovec, Krause, Guestrin, Faloutsos,
  VanBriesen, and Glance]{LeskovecKGFVG07}
Jure Leskovec, Andreas Krause, Carlos Guestrin, Christos Faloutsos, Jeanne
  VanBriesen, and Natalie Glance.
\newblock Cost-effective outbreak detection in networks.
\newblock In \emph{Proceedings of the 13th ACM SIGKDD international conference
  on Knowledge discovery and data mining}, pages 420--429. ACM, 2007.

\bibitem[Lin and Bilmes(2011)]{LinB11}
Hui Lin and Jeff Bilmes.
\newblock A class of submodular functions for document summarization.
\newblock In \emph{Proceedings of the 49th Annual Meeting of the Association
  for Computational Linguistics: Human Language Technologies-Volume 1}, pages
  510--520. Association for Computational Linguistics, 2011.

\bibitem[Lindgren et~al.(2016)Lindgren, Wu, and Dimakis]{LindgrenWD16}
Erik~M. Lindgren, Shanshan Wu, and Alexandros~G. Dimakis.
\newblock Leveraging sparsity for efficient submodular data summarization.
\newblock In \emph{Advances in Neural Information Processing Systems 29: Annual
  Conference on Neural Information Processing Systems}, pages 3414--3422, 2016.

\bibitem[Nemhauser et~al.(1978)Nemhauser, Wolsey, and
  Fisher]{nemhauser1978analysis}
George~L Nemhauser, Laurence~A Wolsey, and Marshall~L Fisher.
\newblock An analysis of approximations for maximizing submodular set
  functions—i.
\newblock \emph{Mathematical programming}, 14:\penalty0 265--294, 1978.

\bibitem[Parambath et~al.(2018)Parambath, Vijayakumar, and
  Chawla]{parambath2018saga}
Shameem~Puthiya Parambath, Nishant Vijayakumar, and Sanjay Chawla.
\newblock Saga: A submodular greedy algorithm for group recommendation.
\newblock In \emph{Proceedings of the AAAI Conference on Artificial
  Intelligence}, volume~32, 2018.

\bibitem[Peng(2021)]{peng2021dynamic}
Binghui Peng.
\newblock Dynamic influence maximization.
\newblock \emph{Advances in Neural Information Processing Systems},
  34:\penalty0 10718--10731, 2021.

\bibitem[Singla et~al.(2016)Singla, Tschiatschek, and Krause]{singla2016noisy}
Adish Singla, Sebastian Tschiatschek, and Andreas Krause.
\newblock Noisy submodular maximization via adaptive sampling with applications
  to crowdsourced image collection summarization.
\newblock In \emph{Proceedings of the AAAI Conference on Artificial
  Intelligence}, volume~30, 2016.

\bibitem[Soare et~al.(2014)Soare, Lazaric, and Munos]{soare2014best}
Marta Soare, Alessandro Lazaric, and R{\'e}mi Munos.
\newblock Best-arm identification in linear bandits.
\newblock \emph{Advances in Neural Information Processing Systems}, 27, 2014.

\bibitem[Takemori et~al.(2020)Takemori, Sato, Sonoda, Singh, and
  Ohkuma]{takemori2020submodular}
Sho Takemori, Masahiro Sato, Takashi Sonoda, Janmajay Singh, and Tomoko Ohkuma.
\newblock Submodular bandit problem under multiple constraints.
\newblock In \emph{Conference on Uncertainty in Artificial Intelligence}, pages
  191--200. PMLR, 2020.

\bibitem[Tschiatschek et~al.(2014)Tschiatschek, Iyer, Wei, and
  Bilmes]{tschiatschek2014learning}
Sebastian Tschiatschek, Rishabh~K Iyer, Haochen Wei, and Jeff~A Bilmes.
\newblock Learning mixtures of submodular functions for image collection
  summarization.
\newblock \emph{Advances in neural information processing systems}, 27, 2014.

\bibitem[Xu et~al.(2018)Xu, Honda, and Sugiyama]{xu2018fully}
Liyuan Xu, Junya Honda, and Masashi Sugiyama.
\newblock A fully adaptive algorithm for pure exploration in linear bandits.
\newblock In \emph{International Conference on Artificial Intelligence and
  Statistics}, pages 843--851. PMLR, 2018.

\bibitem[Yu et~al.(2016)Yu, Fang, and Tao]{yu2016linear}
Baosheng Yu, Meng Fang, and Dacheng Tao.
\newblock Linear submodular bandits with a knapsack constraint.
\newblock In \emph{Proceedings of the AAAI Conference on Artificial
  Intelligence}, volume~30, 2016.

\bibitem[Yue and Guestrin(2011)]{yue2011linear}
Yisong Yue and Carlos Guestrin.
\newblock Linear submodular bandits and their application to diversified
  retrieval.
\newblock \emph{Advances in Neural Information Processing Systems}, 24, 2011.

\end{thebibliography}

\appendix
\appendixtitle{Appendix}

\section{Appendix for Section \ref{sec:adapt}}
In this portion of the appendix, we present missing details and proofs from Section \ref{sec:adapt} in the main paper. We first present Lemma \ref{lem:greedy}, which guarantees that for any standard greedy-based algorithm which satisfies that the element added at each round has the highest marginal gain within an $\epsilon/\kappa$-additive error, then the ultimate solution set satisfies that $f(S)\geq (1-e^{-1})f(OPT)-\epsilon$. Then we propose and analyze the static allocation strategy \algstat in Section \ref{appdx:static}. The pseudocode of \algstat is presented in Algorithm \ref{alg:static} in Section \ref{appdx:static}. Finally, we present the missing proofs and details about the theoretical guarantee of \alg from Section \ref{sec:adapt} in Section \ref{appdx:adapt}. Lemma \ref{lem:greedy} is given as follows.

\begin{lemma}
\label{lem:greedy}
    For any standard greedy-based algorithm proposed for our problem, if with probability at least $1-\delta$, the element $a$ added to the solution set $S$ at each round satisfies that $\Delta f(S,a)\geq\max_{s\in U}\Delta f(S,s)-\epsilon/\kappa$, then it follows that the output solution set $S$ satisfies that with probability at least $1-\delta$
    \begin{align*}
        f(S)\geq(1-e^{-1})f(OPT)-\epsilon.
    \end{align*}
\end{lemma}
\begin{proof}
    The proof of this lemma is similar to the standard greedy algorithm and can be derived from the proof of ExpGreedy in \cite{singla2016noisy}, so we omit the proof here.
\end{proof}
\subsection{Warm-up: static allocation strategy}
\label{appdx:static}
In this section, we present the warm-up algorithm \algstat, which combines the standard greedy algorithm with a subroutine inspired by the static allocation proposed in best-arm identification problem in linear bandit \cite{soare2014best}. The strategy is static in that the sequence of sampled feature vectors $\vect{x}_{a_t}$ is independent from the sampling results. However, unlike the static allocation algorithm proposed in \cite{soare2014best}, which finds the exact optimal solution, here in \algstat, the algorithm finds the element with the highest marginal gain within an $\epsilon/\kappa$ additive error. The theoretical guarantee is stated below. In particular, we first give a description of the static strategy. Next, we provide the detailed statement and proofs of Theorem \ref{thm:static}.

 The algorithm proceeds in rounds. In each round, the algorithm selects the element with the highest marginal gain by taking noisy samples of the marginal gains of adding new elements to the current solution set. For notation simplicity, in the following part, we use $\vect{F}_x$ to denote the marginal gain of adding element $x$ to the solution set in the current round $S$. i.e., $\vect{F}_x=\Delta\vect{F}(S,x)$.  The sampled element is chosen in a greedy manner to make $\max_{x,x'\in U}||\vect{F}_x-\vect{F}_{x'}||_{\vect{A}_t^{-1}}$ to decrease as fast as possible. It is worthy to note that in \algstat, the update of parameters $\hat{\vect{w}}_t$ is calculated based on the least-square estimator. Therefore,  $\vect{A}_{t+1}=\vect{A}_t+\vect{x}_{a_t}\vect{x}_{a_t}^T$, $\vect{b}_{t+1}=\vect{b}_{t}+r_t\vect{x}_{a_t}$, and that $\hat{\vect{w}}_{t+1}=\vect{A}^{-1}_{t+1}\vect{b}_{t+1}$. The parameter $D_t$ is defined as $D_t=R\sqrt{2\log(\frac{\pi^2t^2\kappa n^2}{3\delta})}$.

\begin{algorithm}[t]
\caption{\alglongstat (\algstat)}\label{alg:static}
 \begin{algorithmic}[1]
\STATE \textbf{Input:} $\delta, \epsilon$
 \STATE \textbf{Output: } solution set $S\subseteq U$
 \STATE  $S\gets \emptyset$
 \STATE  $\vect{A}\gets\vect{0}_{d\times d}$, $\vect{b}\gets\vect{0}_{d}$
 \FOR{$l=1,...,\kappa$}
 \STATE $t\gets1$
 \STATE Define $\vect{F}_x=\Delta \vect{F}(S,x)$
 \WHILE{ \textbf{true}}
 
 \STATE
\label{line:find_xt}$x_t=\arg\min_{x\in U }\max_{x',x''\in U }(\vect{F}_{x'}-\vect{F}_{x''})^T(\vect{A}_{t}+\vect{F}_{x}\vect{F}_{x}^T)^{-1}(\vect{F}_{x'}-\vect{F}_{x''})$
 \STATE Obtain the reward $r_t$ of sampling $\vect{F}_{x_t}$
 \STATE \label{line:update_at}
 Update $\vect{A}_{t+1}=\vect{A}_t+\vect{F}_{x_t}\vect{F}_{x_t}^T$, $\vect{b}_{t+1}=\vect{b}_{t}+r_t\vect{F}_{x_t}$
 \STATE Update $\hat{\vect{w}}_{t+1}=\vect{A}^{-1}_{t+1}\vect{b}_{t+1}$
 \IF{ $\exists x\in U $, $\forall x'\in U $, $D_t||\vect{F}_x-\vect{F}_{x'}||_{\vect{A}_t^{-1}}\leq (\vect{F}_x- \vect{F}_{x'})^T\hat{\vect{w}}_t+\epsilon$}
  \STATE \textbf{return} $x$
 \ENDIF
  \STATE $t\gets t+1$
 \ENDWHILE
 \ENDFOR
 \end{algorithmic}
\end{algorithm}

 \subsubsection{proof of Theorem \ref{thm:static}}

We now present the missing theoretical proofs for Theorem \ref{thm:static}, where we provide bounds on the expectation of the reward of the output element and the sample complexity of \algstat in Algorithm \ref{alg:static}. 

\begin{theorem}
\label{thm:static}
    With probability at least $1-\delta$, the following statements hold: 
     \begin{enumerate}
        \item The exact function value of the output solution set $S$ satisfies that $f(S)\geq(1-e^{-1})f(OPT)-\epsilon.$ Here $OPT$ is an optimal solution to the SM problem;
         
        \item During each round in \algstat, the number of samples required is upper bounded by $$t\leq\max\{\frac{256R^2d}{\max\{\Delta_{\min}^2,\epsilon^2/\kappa^2\}}\log(\frac{256R^2d\pi n}{\max\{\Delta_{\min}^2,\epsilon^2/\kappa^2\}}\sqrt{\frac{\kappa}{3\delta}}),\frac{d+d^2+2}{2}\} $$
    where $\Delta_{\min}=\min_{\{x'\in [n],x'\neq x^*\}}\vect{F}_{x^*}^T\vect{w}-\vect{F}_{x'}^T\vect{w}$ and $x^*=\arg\max_{x\in[n]}\vect{F}_{x}^T\vect{w}$.
    \end{enumerate}
\end{theorem}
From Lemma \ref{lem:greedy}, we can see that the theorem can be proved immediately by Lemma \ref{lem:static}, which is presented below.
\begin{lemma}
\label{lem:static}
    With probability at least $1-\delta$, in each round of the Algorithm \ref{alg:static}, the selected element $x$ satisfies
    \begin{align}
    \label{eqn:top_mar_gain}
        \vect{F}^T_x\vect{w}\geq \vect{F}^T_{x^*}\vect{w}-\epsilon/\kappa. 
    \end{align}
    Besides, the number of samples required by each round is upper bounded by $$t\leq\max\{\frac{256R^2d}{\max\{\Delta_{\min}^2,\epsilon^2/\kappa ^2\}}\log(\frac{256R^2d\pi n}{\max\{\Delta_{\min}^2,\epsilon^2/\kappa ^2\}}\sqrt{\frac{\kappa}{3\delta}}),\frac{d+d^2+2}{2}\} $$
    where $\Delta_{\min}=\min_{\{x'\in [n],x'\neq x^*\}}\vect{F}_{x^*}^T\vect{w}-\vect{F}_{x'}^T\vect{w}$, and $x^*=\arg\max_{x'\in[n]}\vect{F}_{x'}^T\vect{w}$.
\end{lemma} 
Next, we present the proof of Lemma \ref{thm:greedy}.
\begin{proof}
Our argument to prove Lemma \ref{lem:static} follows a similar approach as \cite{soare2014best}.   First, we provide proof of (\ref{eqn:top_mar_gain}) on the expectation of the reward of the returned element. Notice that by Proposition \ref{prop:staticy}, and by applying a union bound over all $\kappa$ rounds, with probability at least $1-\delta$, we have that for each round $l$, and for any $x,x'\in  U $,
    \begin{align}
    \label{eqn:high_prob}
         |( \vect{F}_{x'}-\vect{F}_x)^T(\vect{w}-\hat{\vect{w}}_t)|\leq || \vect{F}_{x'}-\vect{F}_x||_{\vect{A}_t^{-1}}D_t.
    \end{align} 
    From the stopping condition of the algorithm, we know that when the algorithm stops,  there exists an element $\vect{x}$ such that $\forall \vect{x}'\in  U $
    \begin{align*}
        D_t||\vect{F}_x-\vect{F}_{x'}||_{\vect{A}_t^{-1}}\leq (\vect{F}_x- \vect{F}_{x'})^T\hat{\vect{w}}_t+\epsilon/\kappa.  
    \end{align*}
    Thus 
    \begin{align*}
        (\vect{F}_x-\vect{F}_{x'})^T\vect{w}&=(\vect{F}_x-\vect{F}_{x'})^T(\vect{w}-\hat{\vect{w}}_t)+(\vect{F}_{x}-\vect{F}_{x'})^T\hat{\vect{w}}_t\\
        &\geq -D_t||\vect{F}_x-\vect{F}_{x'}||_{\vect{A}_t^{-1}}+D_t||\vect{F}_x-\vect{F}_{x'}||_{\vect{A}_t^{-1}}-\epsilon/\kappa \geq-\epsilon/\kappa. 
    \end{align*}
    Since the above inequality holds for any element $\vect{x}'\in  U $, we can conclude the proof. In the following part, we prove the sample complexity for this algorithm. By the stopping condition in \algstat in Algorithm \ref{alg:static}, we have that during the execution of the algorithm \algstat,  for any element $x$, there exists at least one element $x'$ such that
    \begin{align*}
        D_t||\vect{F}_x-\vect{F}_{x'}||_{\vect{A}_t^{-1}}\geq (\vect{F}_x- \vect{F}_{x'})^T\hat{\vect{w}}_t+\epsilon/\kappa.  
    \end{align*}
    Therefore, combining the above inequality with (\ref{eqn:high_prob}), it follows that
    \begin{align*}
     (\vect{F}_x-\vect{F}_{x'})^T\vect{w}&=(\vect{F}_x-\vect{F}_{x'})^T(\vect{w}-\hat{\vect{w}}_t)+(\vect{F}_{x}-\vect{F}_{x'})^T\hat{\vect{w}}_t\\
     &\leq 2D_t||\vect{F}_x-\vect{F}_{x'}||_{\vect{A}_t^{-1}}-\epsilon/\kappa.  
    \end{align*}
    Since the above result holds for any $\vect{x}$, we can set $\vect{x}$ to be the optimal solution $\vect{x}^*$, then we would have that 
    \begin{align*}
     (\vect{F}_{x^*}-\vect{F}_{x'})^T\vect{w}\leq 2D_t||\vect{F}_{x^*}-\vect{F}_{x'}||_{\vect{A}_t^{-1}}-\epsilon/\kappa 
    \end{align*}
    for some $x'$. Therefore, \begin{align}
    \label{eqn:sam_complx}
    \Delta_{\min}\leq 2D_t||\vect{F}_{x^*}-\vect{F}_{x'}||_{\vect{A}_t^{-1}}-\epsilon/\kappa \leq 2D_t\max_{x,x'\in U}||\vect{F}_x-\vect{F}_{x'}||_{\vect{A}_t^{-1}}-\epsilon/\kappa 
    \end{align}
    where $\Delta_{\min}=\min_{\{x'\in [n],x'\neq a^*\}}\vect{F}_{x^*}^T\vect{w}-\vect{F}_{x'}^T\vect{w}$.
Notice that by Lemma 7 and the efficient rounding procedure in \cite{soare2014best}, it follows that
\begin{align*}
     \max_{x,x'\in U}||\vect{F}_x-\vect{F}_{x'}||_{\vect{A}_t^{-1}}\leq 2\sqrt{\frac{(1+\frac{d+d^2+2}{2t})d}{t}}.
\end{align*}

Combine this inequality with (\ref{eqn:sam_complx}), we would get
\begin{align*}
    4D_t\sqrt{\frac{(1+\frac{d+d^2+2}{2t})d}{t}}\geq\Delta_{\min}+\epsilon/\kappa. 
\end{align*}
It follows that
\begin{align*}
    t\leq\frac{16(1+\frac{d+d^2+2}{2t})dD_t^2}{(\Delta_{\min}+\epsilon/\kappa )^2}\leq\frac{16(1+\frac{d+d^2+2}{2t})dD_t^2}{\max\{\Delta_{\min}^2,\epsilon^2/\kappa ^2\}}
\end{align*}
We then have that
\begin{align*}
    t\leq\max\{\frac{32dD_t^2}{\max\{\Delta_{\min}^2,\epsilon^2/\kappa ^2\}},\frac{d+d^2+2}{2}\}.
\end{align*}
From the definition of $D_t$, we can get that
\begin{align*}
    t\leq\max\{\frac{64R^2d\log(\frac{n^2t^2\pi^2}{3\delta})}{\max\{\Delta_{\min}^2,\epsilon^2/\kappa ^2\}},\frac{d+d^2+2}{2}\}.
\end{align*}
From Lemma \ref{lem:logx_over_x}, we conclude that
\begin{align*}
    t\leq\max\{\frac{256R^2d}{\max\{\Delta_{\min}^2,\epsilon^2/\kappa ^2\}}\log(\frac{256R^2d\pi n}{\max\{\Delta_{\min}^2,\epsilon^2/\kappa ^2\}}\sqrt{\frac{\kappa}{3\delta}}),\frac{d+d^2+2}{2}\}. 
\end{align*}
\end{proof}

\subsection{Additional Content to Section \ref{sec:adapt}}
\label{appdx:adapt}
In this appendix, we present the missing content in Section \ref{sec:adapt}. First, we provide a detailed discussion on the sampling allocation ratio $p^*$ used in \alg in Section \ref{appdx:discuss_p^*_greedy}. Next, we delve into the proof of Theorem \ref{thm:greedy}, which provides the theoretical guarantee concerning the algorithm \alg.
\subsubsection{Discussion on the sample allocation ratio}\label{appdx:discuss_p^*_greedy}
In our algorithm \alg, we use the allocation ratio $p^*(l',a|l,i,j)$. $p^*(l',a|l,i,j)$ is the asymptotic-optimal ratio of sampling $\Delta \vect{F}(S_{l'},a)$, which is the marginal gain of adding element $a$ to set $S_{l'}$ for decreasing the confidence interval $||\Delta \vect{F}(S,i)-\Delta \vect{F}(S,j)||_{\vect{A}_t^{-1}}$ when the number of samples goes to infinity at the round $l$. By Lemma \ref{lem:opt_equal}, it satisfies that 
\begin{align}
\label{eqn:define_p*}
    p^*(l',a|l,i,j)=\frac{|w^*_{l',a}(i,j)|}{\sum_{l''=1}^l\sum_{a'=1}^n|w^*_{l'',a'}(i,j)|},
\end{align}
 where $\{w^*_{l',a}(i,j)\}_{a\in[n]}$ is the optimal solution of the linear program in (\ref{eqn:optimization}) below 
\begin{align}
\label{eqn:optimization}
    \arg\min_{\{w_{l',a}(i,j)\}}\sum_{l'=1}^l\sum_{a=1}^n|w_{l',a}(i,j)| \qquad s.t.\qquad \Delta \vect{F}(S,i)-\Delta \vect{F}(S,j)=\sum_{l'=1}^l\sum_{a=1}^nw_{l',a}(i,j)\Delta\vect{F}(S_{l'},a).
\end{align}
   Besides, as is illustrated in Section \ref{sec:adapt}
 and in the proof of Theorem \ref{thm:greedy} in Section \ref{appdx:proof_of_adapt}, if we replace the asymptotic allocation ratio $p^*(l',a|l,i,j)$ with any allocation policy $p(l',a|l,i,j)$, we can still obtain a theoretical guarantee on the upper bound of the sample complexity for our algorithm \alg. In particular, by setting the allocation ratio to $p(l',a|l,i,j)=0.5$ for $l'=l$ and $a=i$ or $j$, and $p(l',a|l,i,j)=0$ otherwise, the allocation strategy in \alg is similar to that used in the CLUCB algorithm in \cite{chen2014combinatorial}. Both these two algorithms sample the direction between $\Delta \vect{F}(S,i)$ and $\Delta \vect{F}(S,j)$ to decrease 
 $||\Delta \vect{F}(S,i)-\Delta \vect{F}(S,j)||_{\vect{A}_t^{-1}}$. Although this allocation ratio is not asymptotically optimal and follows a similar principle to the CLUCB algorithm, there are notable differences: (i) Our algorithm \alg leverages the linear structure, and the concentration inequality used is derived from the linear bandit literature. (ii). The linear submodular bandit algorithm allows us to exploit sampling results from past greedy rounds while in the CLUCB algorithm developed for the general bandit setting, we are not allowed to do that.
 In our experiments, we also adopt this allocation ratio.

\subsubsection{Proof of Theorem \ref{thm:greedy}}
\label{appdx:proof_of_adapt}
 In this section, we prove Theorem \ref{thm:greedy}, which guarantees the sample complexity and approximation ratio of \alg. 
From Lemma \ref{lem:greedy}, we can see that proving Theorem \ref{thm:greedy} is equivalent to proving the following result.
 
\begin{lemma}
\label{thm:adapt}
   With probability at least $1-\delta$, in each round of the \alg in Algorithm \ref{alg:LG}, the selected element $a$ satisfies
    \begin{align*}
        \Delta\vect{F}^T(S,a)\vect{w}\geq \max_{x\in [n]}\Delta\vect{F}^T(S,x))\vect{w}-\epsilon/\kappa 
    \end{align*}
      Besides, assume $\lambda\leq\frac{2R^2}{S^2}\log\frac{1}{\delta}$, then during the round $l$, the total number of samples $\tau$ satisfies
        \begin{align*}
    \tau&\leq\sum_{l'\in[l],a\in[n]}\max\{4H^{(l)}_{l',a} R^2(2\log\frac{1}{\delta}+d\log M^{(l)})+1-N_{l',a}^{(l)},0\},
\end{align*}
where $M^{(l)}=\frac{16H_\epsilon^{(l)} R^2L^2}{\lambda}\log(\frac{1}{\delta})+\frac{32(H_\epsilon^{(l)})^2 R^4L^4}{\lambda^2 }+\frac{2(N^{(l)}+nl)L^2}{\lambda d}+2$. Here $H_{l',a}^{(l)}=\max_{i,j\in[n]}\frac{p^*(l',a|l,i,j)\rho_{i,j}^l}{\max\{\epsilon/\kappa,\frac{\Delta_{l,i}\vee\Delta_{l,j}+\epsilon/\kappa}{3}\}^2}$ and $H_{\epsilon}^{(l)}=\sum_{l'\in[l],a\in[n]}H_{l',a}^{(l)}$, where $\rho_{i,j}^l$ is the optimal value of (\ref{eqn:optimization}).
\end{lemma}

\begin{proof}
   Throughout the proof, we assume the clean event $\mathcal{E}$ occurs. The event $\mathcal{E}$ is defined as 
\begin{align*}
    \mathcal{E}=\{\forall t>0,\forall i,j\in[n], |(\Delta\vect{F}(S,i)-\Delta\vect{F}(S,j))^T(\hat{\vect{w}}-\vect{w})|\leq \beta_t(i,j)\},
\end{align*}
where the confidence interval $\beta_t(i,j)=C_t||\Delta\vect{F}(S,i)-\Delta\vect{F}(S,j)||_{\vect{A}_t}^{-1} $ and that $C_t=R\sqrt{2\log\frac{\det(\vect{A}_t)^{\frac{1}{2}}\det(\lambda I)^{-\frac{1}{2}}}{\delta}}+\lambda^{\frac{1}{2}}S$.  Then by Proposition \ref{prop:adaptive}, and the fact that for all $i$, $\Delta\vect{F}(S,i)$ only depends on the past samples of marginal gains, we have the following lemma.
\begin{lemma}
\label{lem:clean_event}
    With probability at least $1-\delta$, event $\mathcal{E}$ holds.
\end{lemma}

First of all, we prove the result on the correctness of the output element $i_{t}$. Let us denote that $a^*=\arg\max_{a\in[n]}\Delta \vect{F}(S,a)^T\vect{w}$, and that $\epsilon'=\epsilon/\kappa$. By the stopping condition in \alg, we have that 
\begin{align*}
    B(t)=(\Delta\vect{F}(S,j_t)-\Delta\vect{F}(S,i_t))^T\hat{\vect{w}}+\beta_{t}(i_t,j_t)\leq\epsilon'.
\end{align*}
By the selection strategy of $j_t$, we have that $B(t)\geq (\Delta\vect{F}(S,a^*)-\Delta\vect{F}(S,i_t))^T\hat{\vect{w}}+\beta_{t}(a^*,i_t)^T$. Therefore, by Lemma \ref{lem:clean_event}, we have that
\begin{align*}
    (\Delta\vect{F}(S,a^*)-\Delta\vect{F}(S,i_t))^T\vect{w}\leq (\Delta\vect{F}(S,a^*)-\Delta\vect{F}(S,i_t))^T\hat{\vect{w}}+\beta_{t}(a^*,i_t)\leq \epsilon'.
\end{align*}
Next, we prove the theoretical guarantee of the sample complexity. Let us denote  
$\Delta_{l,i}$ to be the gap between element $i$ and the element with the highest marginal gain at round $l$, i.e., $\Delta_{l,i}=\max_{x\in[n]}\Delta\vect{F}(S_l,x)^T\vect{w}-\Delta\vect{F}(S_l,i)^T\vect{w}$. At the round $l$, we have that the solution set $S=S_l$. For notation simplicity, we denote $\Delta_{i}=\max_{x\in[n]}\Delta\vect{F}(S,x)^T\vect{w}-\Delta\vect{F}(S,i)^T\vect{w}$. From Lemma 4 in the appendix of \cite{xu2018fully}, we have
\begin{lemma}
\label{lem:upperBound_Bt}
    Under event $\mathcal{E}$, throughout the execution of \alg, $B(t)$ can be bounded as 
    \begin{align*}
        B(t)\leq\min(0,-(\Delta_{i_t}\vee\Delta_{j_t})+2\beta_t(i_t,j_t))+\beta_t(i_t,j_t). 
    \end{align*}
\end{lemma}
 The proof is the same as in \cite{xu2018fully}, so we omit the proof here. First of all, we derive the bound on the total number of samples for each element $a$ from the sampling set $\mathcal{X}$.  Then by Lemma \ref{lem:upperBound_Bt} and that $B(t)\geq\epsilon'$ we have that 
\begin{align*}
    \beta_t(i_t,j_t)\geq\max\{\epsilon',\frac{\Delta_{i_t}\vee\Delta_{j_t}+\epsilon'}{3}\}.
\end{align*}
By the definition of the confidence interval $\beta_t(i_t,j_t)$, we have that
\begin{align*}
    ||\Delta\vect{F}(S,i_t)-\Delta\vect{F}(S,j_t)||_{\vect{A}_t^{-1}}\geq\frac{\max\{\epsilon',\frac{\Delta_{i_t}\vee\Delta_{j_t}+\epsilon'}{3}\}}{C_t}.
\end{align*}
Since we have 
\begin{align}
\label{eqn:bound_norm_difference}
    &||\Delta\vect{F}(S,i_t)-\Delta\vect{F}(S,j_t)||_{\vect{A}_t^{-1}}\\=&\sqrt{(\Delta\vect{F}(S,i_t)-\Delta\vect{F}(S,j_t))^T\vect{A}_{t}^{-1}(\Delta\vect{F}(S,i_t)-\Delta\vect{F}(S,j_t))}\nonumber
    \\=&\sqrt{(\Delta\vect{F}(S,i_t)-\Delta\vect{F}(S,j_t))^T(\sum_{l''\in[l]}\sum_{a'\in[n]}T_{l'',a'}(t)\Delta \vect{F}(S_{l''},a')\Delta \vect{F}(S_{l''},a')^T))^{-1}(\Delta\vect{F}(S,i_t)-\Delta\vect{F}(S,j_t))}.
\end{align}

Suppose the marginal gain of adding an element $a$ to subset $S_{l'}$ is sampled at time $t$, from the element selection strategy, we have that for any other past marginal gain vectors, it follows that for any $l''\in[l]$ and $a'\in[n]$, 
\begin{align*}
    \frac{ T_{l',a}(t)}{p^*(l',a|l,i_t,j_t)}\leq \frac{ T_{l'',a'}(t)}{p^*(l'',a'|l,i_t,j_t)}.
\end{align*}
Combine (\ref{eqn:bound_norm_difference}) with the above inequality, it follows that 
\begin{align*}
     ||\Delta\vect{F}(S,i_t)-\Delta\vect{F}(S,j_t)||_{\vect{A}_t^{-1}}\leq\sqrt{\frac{p^*(l',a|l,i_t,j_t)}{T_{l',a}(t)}}||\Delta\vect{F}(S,i_t)-\Delta\vect{F}(S,j_t)||_{\Lambda_{i_t,j_t}^{-1}}
\end{align*}
where we define 
\begin{align}
    \label{eqn:define_lambda}\Lambda_{i,j}=\sum_{l''\in[l]}\sum_{a'\in[n]}p^*(l'',a'|l,i,j)\Delta \vect{F}(S_{l''},a')\Delta \vect{F}^T(S_{l''},a'). 
\end{align}
Therefore, we have 
\begin{align*}
   T_{l',a}(t)\leq\frac{p^*(l',a|l,i_t,j_t)||\Delta\vect{F}(S,i_t)-\Delta\vect{F}(S,j_t)||^2_{\Lambda_{i_t,j_t}^{-1}}C_t^2}{\max\{\epsilon',\frac{\Delta_{i_t}\vee\Delta_{j_t}+\epsilon'}{3}\}^2} \\
   \leq\max_{i,j\in[n]}\frac{p^*(l',a|l,i,j)||\Delta\vect{F}(S,i)-\Delta\vect{F}(S,j)||^2_{\Lambda_{i,j}^{-1}}C_t^2}{\max\{\epsilon',\frac{\Delta_{i}\vee\Delta_{j}+\epsilon'}{3}\}^2} 
\end{align*}
Let $\tau'$ denote the total number of samples in round $l$, encompassing both the offline dataset and the samples acquired during the execution of \alg. Consequently, upon termination of the algorithm, the cumulative number of samples associated with element $a$ is bounded by:
\begin{align} \label{eqn:total_samples}
T_{l',a}(\tau')\leq\max_{i,j\in[n]}\frac{p^*(l',a|l,i,j)||\Delta\vect{F}(S,i)-\Delta\vect{F}(S,j)||^2_{\Lambda_{i,j}^{-1}}C_{\tau'}^2}{\max\{\epsilon',\frac{\Delta_{i}\vee\Delta_{j}+\epsilon'}{3}\}^2}+1.
\end{align}
Notice that we can reuse $N_{l',a}^{(l)}$ samples from the past rounds, therefore, the total number of samples of element $a$ in round $l$ in \alg is bounded by 
\begin{align*}
    \max\{T_{l',a}(\tau')-N_{l',a}^{(l)},0\}.
\end{align*}
The total number of samples for all the elements in round $l$ during the execution of \alg is bounded by
\begin{align*}
    \tau\leq \sum_{l'\in[l]}\sum_{a\in[n]}\max\{T_{l',a}(\tau')-N_{l',a}^{(l)},0\}.
\end{align*}
 Combine this inequality with (\ref{eqn:total_samples}), we get that
\begin{align*}
    \tau&\leq\sum_{l'\in[l]}\sum_{a\in[n]}\max\{\max_{i,j\in[n]}\frac{p^*(l',a|l,i,j)||\Delta\vect{F}(S,i)-\Delta\vect{F}(S,j)||^2_{\Lambda_{i,j}^{-1}}C_{\tau'}^2}{\max\{\epsilon',\frac{\Delta_{i}\vee\Delta_{j}+\epsilon'}{3}\}^2}+1-N_{l',a}^{(l)},0\}
   \\&\leq\sum_{l'\in[l]}\sum_{a\in[n]}\max\{\max_{i,j\in[n]}\frac{p^*(l',a|l,i,j)||\Delta\vect{F}(S,i)-\Delta\vect{F}(S,j)||^2_{\Lambda_{i,j}^{-1}}C_{\tau+N^{(l)}}^2}{\max\{\epsilon',\frac{\Delta_{i}\vee\Delta_{j}+\epsilon'}{3}\}^2}+1-N_{l',a}^{(l)},0\}
\end{align*}
where $N^{(l)}=\sum_{l'\in[l]}\sum_{a\in[n]}N_{l',a}^{(l)}$ is the total number of samples in the offline dataset. 

 From Lemma \ref{lem:opt_equal} and the discussion given in Appendix B in \cite{xu2018fully}, we have that the allocation ratio $p^*$ is the optimal solution of the optimization problem as defined in (\ref{eqn:define_p*}) and (\ref{eqn:optimization}), 
then it follows that 
\begin{align*}
    ||\Delta\vect{F}(S,i)-\Delta\vect{F}(S,j)||^2_{\Lambda_{i,j}^{-1}}\leq \rho_{i,j}^l.
\end{align*}
where $\rho_{i,j}^l$ is the optimal value of (\ref{eqn:optimization}) in the round $l$.

Let us denote $H_{l',a}^{(l)}=\max_{i,j\in[n]}\frac{p^*(l',a|l,i,j)\rho_{i,j}^l}{\max\{\epsilon/\kappa,\frac{\Delta_{l,i}\vee\Delta_{l,j}+\epsilon/\kappa}{3}\}^2}$ and $H_{\epsilon}^{(l)}=\sum_{l'\in[l],a\in[n]}H_{l',a}^{(l)}$. Then the inequality above is 
\begin{align}
    \label{eqn:ineq_on_tau'}\tau&\leq\sum_{l'\in[l]}\sum_{a\in[n]}\max\{\Ha C_{\tau+N}^2+1-N_{l',a}^{(l)},0\}.
\end{align}
It then follows that 
\begin{align}
\label{eqn:rough_ineq_on_tau'}
    \tau&\leq H_{\epsilon}^{(l)} C_{\tau+N}^2+nl.
\end{align}
By definition, we have that 
\begin{align*}
    C_t=R\sqrt{2\log\frac{\det(\vect{A}_t)^{\frac{1}{2}}\det(\lambda I)^{-\frac{1}{2}}}{\delta}}+\lambda^{\frac{1}{2}}S.
\end{align*}
By Lemma \ref{lem:bound_det_A}, we can get that 
\begin{align*}
C_t\leq R\sqrt{2\log\frac{1}{\delta}+d\log(1+\frac{tL^2}{\lambda d})}+\lambda^{\frac{1}{2}}S.
\end{align*}
Since $\lambda\leq\frac{2R^2}{S^2}\log\frac{1}{\delta}$, it follows that 
\begin{align}
\label{eqn:bound_C_t}
   C_t\leq 2R\sqrt{2\log\frac{1}{\delta}+d\log(1+\frac{tL^2}{\lambda d})}.
\end{align}
Combine (\ref{eqn:rough_ineq_on_tau'}) with (\ref{eqn:bound_C_t}), we can get
\begin{align*}
    \tau\leq4H_\epsilon^{(l)}R^2\big(2\log\frac{1}{\delta}+d\log(1+\frac{(\tau+N^{(l)})L^2}{\lambda d})\big)+nl
\end{align*}
It then follows that 
\begin{align*}
    \frac{(\tau+N^{(l)})L^2}{\lambda d}+1\leq\frac{4H_\epsilon^{(l)} R^2L^2}{\lambda d}\big(2\log\frac{1}{\delta}+d\log(1+\frac{(\tau+N^{(l)})L^2}{\lambda d})\big)+\frac{(N^{(l)}+nl)L^2}{\lambda d}+1
\end{align*}
By Lemma \ref{lem:calculation1}, we have that 
\begin{align*}
    \frac{(\tau+N^{(l)})L^2}{\lambda d}+1&\leq\frac{16H_\epsilon^{(l)} R^2L^2}{\lambda d}\log\frac{1}{\delta}+\frac{8H_\epsilon^{(l)} R^2L^2}{\lambda }\log(\frac{4H_\epsilon R^2L^2}{\lambda})+\frac{2(N^{(l)}+nl)L^2}{\lambda d}+2\\
    &\leq \frac{16H_\epsilon^{(l)} R^2L^2}{\lambda}\log\frac{1}{\delta}+\frac{32(H_\epsilon^{(l)})^2 R^4L^4}{\lambda^2 }+\frac{2(N^{(l)}+nl)L^2}{\lambda d}+2.
\end{align*}
 Plugging the above inequality into (\ref{eqn:ineq_on_tau'}), it follows that 

\begin{align*}
    \tau&\leq\sum_{l'\in[l]}\sum_{a\in[n]}\max\{4\Ha R^2(2\log\frac{1}{\delta}+d\log M)+1-N_{l',a}^{(l)},0\}.
\end{align*}
where $M=\frac{16H_\epsilon^{(l)} R^2L^2}{\lambda}\log\frac{1}{\delta}+\frac{32(H_\epsilon^{(l))^2} R^4L^4}{\lambda^2 }+\frac{2(N^{(l)}+nl)L^2}{\lambda d}+2.$ 
\end{proof}

\section{Appendix for Section \ref{sec:threshold}}
\label{appdx:threshold}
In this section, we present the proofs of theoretical results from Section \ref{sec:threshold}. First, in Section \ref{appdx:proof_of_samp}, we present and prove the Lemma \ref{thm:samp} concerning the theoretical guarantee on the value of the evaluated marginal gain and sample complexity of \algthre. Next, we apply the theoretical results of Lemma \ref{thm:samp} and prove the final theoretical guarantee for the linear threshold greedy algorithm \algthre in Algorithm \ref{alg:ATG}.
\subsection{Proof of Lemma \ref{thm:samp}}
\label{appdx:proof_of_samp}
We prove the theoretical guarantee about evaluation of each marginal gain in \algthre. First of all, we present the detailed statements of Lemma \ref{thm:samp}. First, we notice that by Lemma \ref{lem:opt_equal}, we have that $p^*$ at the $m$-th evaluation of marginal gain can be calculated by

\begin{align} \label{eqn:define_p^*_tg}
    p^*_i=\frac{|w^*_i|}{\sum_{i\in[m]}|w^*_i|}
\end{align}
where $w_i^*$ is the optimal solution to the linear programming problem below.
 \begin{align}
    \label{eqn:opt_tg}
        \arg\min_{\{w_{i}\}}\sum_{i\in[m]}|w_{i}| \qquad s.t.\qquad  \Delta\vect{F}(S_m,a_m)=\sum_{i=1}^mw_{i}\Delta\vect{F}(S_i,a_i).
    \end{align}
    In addition, we denote the optimal value of the above optimization problem as $\rho^{(m)}$. In our experiments, we also consider the allocation ratio of $p_i=0$ for $i\neq m$ and $p_m=1$, i.e., the algorithm always samples the marginal gain $\Delta f(S_m,a_m)$ to determine whether  $\Delta f(S_m,a_m)$ is approximately above $w$ or not.
 \begin{lemma}
     \label{thm:samp}
    During the $m$-th evaluation on the marginal gain of adding element $a_m$ to $S_m$ in \algthre, we have that with probability at least $1-\delta/2$, the following statement holds. 
    \begin{enumerate}
        \item Assuming $\lambda\leq\frac{2R^2}{S^2}\log\frac{2}{\delta}$, then it takes at most
    \begin{align*}
        \sum_{i\in[m]}\max\{4H^{(m)}_{i} R^2(2\log\frac{2}{\delta}+d\log M^{(m)})+1-N_{i}^{(m)},0\}
\end{align*}
  samples to evaluate the $m$-th marginal gain, where $M^{(m)}=\frac{16H_\epsilon^{(m)} R^2L^2}{\lambda}\log\frac{2}{\delta}+\frac{32(H_\epsilon^{(m)})^2 R^4L^4}{\lambda^2 }+\frac{2(N^{(m)}+m)L^2}{\lambda d}+2$. 
  Here $H_{i}^{(m)}=\frac{p^*_{i}\rho^{(m)}}
  {\max\{\frac{\epsilon/\kappa+|w-\Delta \vect{F}(S,a)^T\vect{w}|}{2},\epsilon/\kappa\}^2}$ 
  and $H_{\epsilon}^{(m)}=\sum_{i\in[m]}H_{i}^{(m)}$. 
  $p^*_{i}$ is the fraction of samples allocated to sample $\Delta \vect{F}(S_i,a_i)$ and therefore $\sum_{i\in[m]}p^*_{i}=1$. $\rho^{(m)}$ is the optimal value of 
    \begin{align}
    \label{lem:opt}
        \arg\min_{\{w_{i}\}}\sum_{i\in[m]}|w_{i}| \qquad s.t.\qquad  \Delta\vect{F}(S_m,a_m)=\sum_{i=1}^mw_{i}\Delta\vect{F}(S_i,a_i).
    \end{align}
    \item If the element is added to the solution set, then $\Delta\vect{F}(S_m,a_m)^T\vect{w}\geq w-\epsilon/\kappa$. Otherwise,  $\Delta\vect{F}(S_m,a_m)^T\vect{w}\leq w+\epsilon/\kappa$.
    \end{enumerate}
    
 \end{lemma}

For notation simplicity, throughout the proof, we define $\epsilon'=\epsilon/\kappa$. To prove this lemma, We first present the following Lemma \ref{lem:samp_clean_event} and Lemma \ref{lem:conf_int} about our algorithm \algthre.
Once these lemmas are proven, we next prove the main result of Lemma \ref{thm:samp} using the lemmas. We start by introducing the following "clean event".
    \begin{align*}
        \mathcal{E}=\{|\Delta\vect{F}(S_m,a_m)^T(\hat{\vect{w}}-\vect{w})|\leq \beta_t,\forall t\}
    \end{align*}

    Here $\beta_t$ is the confidence interval and is defined as $\beta_t = C_t||\Delta\vect{F}(S_m,a_m)||_{\vect{A}_t^{-1}}=(R\sqrt{2\log\frac{2\det(\vect{A}_t)^{\frac{1}{2}}\det(\lambda I)^{-\frac{1}{2}}}{\delta}}+\lambda^{\frac{1}{2}}S)||\Delta\vect{F}(S_m,a_m)||_{\vect{A}_t^{-1}}$. From the fact that  $\Delta\vect{F}(S_m,a_m)$ only depends on the offline dataset, i.e., the history up to time $t$, we can apply the result in Proposition \ref{prop:adaptive}. It follows that $P(\mathcal{E})\geq 1-\frac{\delta}{2}$. Therefore, we have the following lemma.
\begin{lemma}
\label{lem:samp_clean_event}
    With probability at least $1-\delta/2$, it holds that 
    \begin{align*}
        P(\mathcal{E})\geq 1-\delta.
    \end{align*}
    
\end{lemma}
Next, we prove the bound on the confidence interval $\beta_t$.
\begin{lemma}
\label{lem:conf_int}
    Conditioned on the event $\mathcal{E}$, the confidence interval $\beta_t$ satisfies that 
    \begin{align*}
    \beta_t\geq\max\{\frac{\epsilon'+|w-\Delta\vect{F}(S_m,a_m)^T\vect{w}|}{2},\epsilon'\}
    \end{align*}
    before the evaluation of the marginal gain ends.
\end{lemma}
\begin{proof}
    It is equivalent to prove that when $\beta_t\leq \frac{\epsilon'+|w-\Delta\vect{F}(S_m,a_m)^T\vect{w}|}{2}$, the evaluation of the marginal gain ends. 
       If $\beta_t\leq\frac{\epsilon'+w-\Delta\vect{F}(S_m,a_m)^T\vect{w}}{2}$, then we have $\Delta\vect{F}(S_m,a_m)^T\vect{w}\leq w+\epsilon'-2\beta_t $. From Lemma \ref{lem:clean_event}, we have that with probability at least $1-\delta$, it holds that $\Delta\vect{F}(S_m,a_m)^T\hat{\vect{w}}-\Delta\vect{F}(S_m,a_m)^T\vect{w}\leq\beta_t$. Therefore,
     \begin{align*}
         &\Delta\vect{F}(S_m,a_m)^T\hat{\vect{w}}+\beta_t\\
         &\leq (\Delta\vect{F}(S_m,a_m)^T\hat{\vect{w}}-\Delta\vect{F}(S_m,a_m)^T\vect{w})+\Delta\vect{F}(S_m,a_m)^T\vect{w}+\beta_t\\
         &\leq w+\epsilon'.
     \end{align*}
     Thus the algorithm ends.
     
     Similarly, we consider the case where $\beta_t\leq\frac{\epsilon'-w+\Delta\vect{F}(S_m,a_m)^T\vect{w}}{2}$. In this case, we have that $\Delta\vect{F}(S_m,a_m)^T\vect{w}\geq 2\beta_t+w-\epsilon'$.
    Notice that conditioned on the clean event defined in Lemma \ref{lem:clean_event}, we have that $\Delta\vect{F}(S_m,a_m)^T\hat{\vect{w}}-\Delta\vect{F}(S_m,a_m)^T\vect{w}\geq-\beta_t$. Then
    \begin{align*}
        \Delta\vect{F}(S_m,a_m)^T\hat{\vect{w}}-\beta_t&\geq \Delta\vect{F}(S_m,a_m)^T\hat{\vect{w}}-\Delta\vect{F}(S_m,a_m)^T\vect{w}\\
        &\qquad+\Delta\vect{F}(S_m,a_m)^T\vect{w}-\beta_t\\
        &\geq-\beta_t+2\beta_t\\
        &\qquad+w-\epsilon'-\beta_t\\
        &=w-\epsilon'.
    \end{align*}
    Therefore, the algorithm ends. 
    In the case where $\beta_t\leq \epsilon'$, then $w+\epsilon'-\beta_t\geq w-\epsilon'+\beta_t$. Therefore, either we have $\Delta\vect{F}(S_m,a_m)^T\hat{\vect{w}}\geq w+\beta_t-\epsilon'$ or $\Delta\vect{F}(S_m,a_m)^T\hat{\vect{w}}\leq w-\beta_t+\epsilon'$. Thus the the evaluation of the $m$-th marginal gain ends. 
\end{proof}

Now we can prove Lemma \ref{thm:samp}.

   \begin{proof}
    
First of all, we prove the second result in the Lemma \ref{thm:samp}. Conditioned on the event $\mathcal{E}$ , we have that if the $a_m$ element is added to $S_m$, then
\begin{align*}
\Delta\vect{F}(S_m,a_m)^T\vect{w}\geq\Delta\vect{F}(S_m,a_m)^T\hat{\vect{w}}-\beta_t\geq w-\epsilon'.
\end{align*}
If the $a_m$ element is not added to $S_m$, then 
\begin{align*}
    \Delta\vect{F}(S_m,a_m)^T\vect{w}\leq \Delta\vect{F}(S_m,a_m)^T\hat{\vect{w}}+\beta_t\leq w+\epsilon'.
\end{align*}
Next, we prove our first result, which provides the theoretical guarantee of sample complexity. The proof depends on Lemma \ref{lem:conf_int}. By Lemma \ref{lem:conf_int}, and the definition of $\beta_t$, we have that 
\begin{align*}
    C_t||\Delta\vect{F}(S_m,a_m)||_{\vect{A}_t^{-1}}\geq\max\{\frac{\epsilon'+|w-\Delta\vect{F}(S_m,a_m)^T\vect{w}|}{2},\epsilon'\}.
\end{align*}
Similar to the proof in Lemma \ref{lem:upperBound_Bt}, we have that the sampled element $i_t$ satisfies 
\begin{align*}  ||\Delta\vect{F}(S_m,a_m)||_{\vect{A}_t^{-1}}\leq\sqrt{\frac{p_{i_t}^*}{T_{i_t}(t)}}||\Delta\vect{F}(S_m,a_m)||_{\Lambda^{-1}},
\end{align*}
where $\Lambda=\sum_{i\in[m]}p_i^*\Delta\vect{F}(S_i,a_i)\Delta\vect{F}(S_i,a_i)^T$. Therefore, before the marginal gain of adding $a_{i_t}$ to $S_{i_t}$ is sampled at time $t$, we have 
\begin{align*}
   T_{i_t}(t)\leq\frac{p_{i_t}^*||\Delta\vect{F}(S_m,a_m)||^2_{\Lambda^{-1}}C_t^2}{\max\{\frac{\epsilon'+|w-\Delta\vect{F}(S_m,a_m)^T\vect{w}|}{2},\epsilon'\}^2}.
\end{align*}
For any other marginal gain with index $i$, suppose $\tilde{t}$ is the last time $\Delta \vect{F}(S_i,a_i)$ is sampled, and that $\tau'$ is the total number of samples, then 
\begin{align*}
    T_i(\tau')= T_{i}(\tilde{t})+1\leq\frac{p_{i}^*||\Delta\vect{F}(S_m,a_m)||^2_{\Lambda^{-1}}C_{\tau'}^2}{\max\{\frac{\epsilon'+|w-\Delta\vect{F}(S_m,a_m)^T\vect{w}|}{2},\epsilon'\}^2}+1.
\end{align*}
Since $T_i(\tau')$ is the total number of samples to the marginal gain $\Delta \vect{F}(S_i,a_i)$ including the past samples, then the total number of samples of $\Delta \vect{F}(S_i,a_i)$ during the evaluation of $\Delta \vect{F}(S_m,a_m)$  is upper bounded by
\begin{align*}
   \max\{\frac{p_{i}^*||\Delta\vect{F}(S_m,a_m)||^2_{\Lambda^{-1}}C_{\tau'}^2}{\max\{\frac{\epsilon'+|w-\Delta\vect{F}(S_m,a_m)^T\vect{w}|}{2},\epsilon'\}^2}+1-N_i^{(m)},0\}.
\end{align*}
Summing over all elements, we would get
\begin{align*}
\tau&\leq\sum_{i\in[m]}\max\{\frac{p_{i}^*||\Delta\vect{F}(S_m,a_m)||^2_{\Lambda^{-1}}C_{\tau'}^2}{\max\{\frac{\epsilon'+|w-\Delta\vect{F}(S_m,a_m)^T\vect{w}|}{2},\epsilon'\}^2}+1-N_i^{(m)},0\}\\
&\leq\sum_{i\in[m]}\max\{\frac{p_{i}^*||\Delta\vect{F}(S_m,a_m)||^2_{\Lambda^{-1}}C_{\tau+N^{(m)}}^2}{\max\{\frac{\epsilon'+|w-\Delta\vect{F}(S_m,a_m)^T\vect{w}|}{2},\epsilon'\}^2}+1-N_i^{(m)},0\},
\end{align*}
where $N^{(m)}$ is the total number of samples in the offline dataset. By Lemma \ref{lem:opt_equal}, it follows that 
\begin{align*}
\tau
&\leq\sum_{i\in[m]}\max\{\frac{p_{i}^*\rho^{(m)} C_{\tau+N^{(m)}}^2}{\max\{\frac{\epsilon'+|w-\Delta\vect{F}(S_m,a_m)^T\vect{w}|}{2},\epsilon'\}^2}+1-N_i^{(m)},0\},
\end{align*}
where $\rho^{(m)}$ is the optimal value of the optimization problem in (\ref{eqn:opt_tg}). 
Let us denote $H_{i}^{(m)}=\frac{p_{i}^*\rho^{(m)} }{\max\{\frac{\epsilon'+|w-\Delta\vect{F}(S_m,a_m)^T\vect{w}|}{2},\epsilon'\}^2}$ and $H_{\epsilon}^{(m)}=\sum_{i\in[m]}H_{i}^{(m)}=\frac{\rho^{(m)} }{\max\{\frac{\epsilon'+|w-\Delta\vect{F}(S_m,a_m)^T\vect{w}|}{2},\epsilon'\}^2}$. Then the inequality above is 
\begin{align}
\label{eqn:ineq_on_tau_thre}\tau&\leq\sum_{i\in[m]}\max\{H_i^{(m)} C_{\tau+N^{(m)}}^2+1-N_i^{(m)},0\}.
\end{align}
It then follows that 
\begin{align}
    \tau&\leq H_{\epsilon}^{(m)} C_{\tau+N^{(m)}}^2+m
\end{align}

Similar to the proof of Theorem \ref{thm:adapt}, we have that
\begin{align*}
    \tau&\leq\sum_{i\in[m]}\max\{4H_i^{(m)} R^2(2\log\frac{2}{\delta}+d\log M^{(m)})+1-N_i^{(m)},0\}.
\end{align*}
where $M^{(m)}=\frac{16H_\epsilon^{(m)} R^2L^2}{\lambda}\log\frac{2}{\delta}+\frac{32(H_\epsilon^{(m)})^2 R^4L^4}{\lambda^2 }+\frac{2(N^{(m)}+m)L^2}{\lambda d}+2.$
\end{proof} 

\subsection{Proof of Theorem \ref{thm:threshold}}

In this section, we present the proof of Theorem \ref{thm:threshold}.

\textbf{Theorem \ref{thm:threshold}. }\textit{ \algthre makes $n\log(\kappa/\alpha)/\alpha$ evaluations of the marginal gains. In addition, with probability at least $1-\delta$, the following statements hold: 
     \begin{enumerate}
        \item The function value of the output solution set $S$ satisfies that $f(S)\geq(1-e^{-1}-\alpha)f(OPT)-2\epsilon.$ Here $OPT$ is an optimal solution to the SM problem;
        \item Assuming $\lambda\leq\frac{2R^2}{S^2}\log\frac{2}{\delta}$, the $m$-th evaluation of the marginal gain of adding an element $a$ to $S$ in \algthre takes at most
    \begin{align*}
        \sum_{i\in[m]}\max\{4H^{(m)}_{i} R^2(2\log\frac{2}{\delta}+d\log M^{(m)})+1-N_{i}^{(m)},0\}
\end{align*}
  samples, where $M^{(m)}=\frac{16H_\epsilon^{(m)} R^2L^2}{\lambda}\log\frac{2}{\delta}+\frac{32(H_\epsilon^{(m)})^2 R^4L^4}{\lambda^2 }+\frac{2(N^{(m)}+m)L^2}{\lambda d}+2$. 
  Here $H_{i}^{(m)}=\frac{p^*_{i}\rho^{(m)}}
  {\max\{\frac{\epsilon/\kappa+|w-\Delta \vect{F}(S_m,a_m)^T\vect{w}|}{2},\epsilon/\kappa\}^2}$ 
  and $H_{\epsilon}^{(m)}=\sum_{i\in[m]}H_{i}^{(m)}$. 
  $p^*_{i}$ is the fraction of samples allocated to sample $\Delta \vect{F}(S_i,a_i)$. $N^{(m)}$ is the total number of samples before the $m$-th evaluation of marginal gains. $N^{(m)}_i$ is the number of samples to the marginal gain $\Delta f(S_i,a_i)$ before the $m$-th evaluation of marginal gains.
    \end{enumerate}
    }
The second result can be obtained by applying the result in Lemma \ref{thm:samp}. To prove the first result, we first present a series of needed lemmas. In order for the guarantees of Theorem \ref{thm:threshold} to hold, two random events must occur during \alg. The first event is that the estimate of the max singleton value of $f$ on Line \ref{alg:ATG:line:sample-mean} in \alg is an $\epsilon$-approximation of its true value. More formally, we have the following lemma.
 \begin{lemma}
    \label{lem:clean_event_monotone}
      With probability at least $1-\delta/3$, we have $\max_{s\in U}f(s)-\epsilon\leq g\leq\max_{s\in U}f(s)+\epsilon$.
\end{lemma}
\begin{proof}
    For a fix $s\in U$, by Hoeffding's inequality we would have that
    \begin{align}
        P(|\hat{f}(s)-f(s)|\geq\epsilon)\leq\frac{\delta}{3n}.
    \end{align}
    Taking a union bound over all elements we would have that
    \begin{align*}
        P(\exists s\in U, s.t.|\hat{f}(s)-f(s)|\geq\epsilon )\leq\frac{\delta}{3}.
    \end{align*}
    Then with probability at least $1-\frac{\delta}{3}$, $|\hat{f}(s)-f(s)|\leq\epsilon$ 
    for all $s\in U$.  
    It then follows that $\forall s\in U$, $f(s)-\epsilon\leq\hat{f}(s)\leq f(s)+\epsilon$. Therefore $$\max_{s\in U}(f(s)-\epsilon)\leq\max_{s\in U}\hat{f}(s)\leq\max_{s\in U}(f(s)+\epsilon).$$ Thus we have
    \begin{align*}
        \max_{s\in U}f(s)-\epsilon\leq g\leq\max_{s\in U}f(s)+\epsilon.
    \end{align*} 
    
\end{proof}

With the above Lemma \ref{lem:clean_event_monotone}  and Lemma \ref{thm:samp}, and by taking the union bound, we have that with probability at least $1-\delta$, the two events in Lemma \ref{lem:clean_event_monotone}  and Lemma \ref{thm:samp} both hold during the \alg. Our next step is to show that if both of the events occur during \alg, the approximation guarantees and sample complexity of Theorem \ref{thm:threshold} hold. To this end, we need the following Lemma \ref{lem:mar_gain}. 

\begin{lemma}
\label{lem:mar_gain}
    Assume the events defined in Lemma \ref{lem:clean_event_monotone} and Lemma \ref{thm:samp} above hold during \alg. Then for any element $s$ that is added to the solution set $S$, the following statement holds.
    \begin{align*}
        \Delta f(S,s)\geq \frac{1-\alpha}{\kappa}(f(OPT)-f(S))-2\epsilon.
    \end{align*}
\end{lemma}

\begin{proof}   
    At the first round in the for loop from Line \ref{line:tg_for_loop_starts} to Line \ref{line:tg_for_loop_ends} in Algorithm \ref{alg:ATG}, if an element $s$ is added to the solution set, it holds by Lemma \ref{lem:clean_event_monotone} that $\Delta f(S,s)\geq w-\epsilon$.
   Since at the first round $w=g$ and $g\geq \max_{s\in U}f(s)-\epsilon$. It follows that $\Delta f(S,s)\geq \max_{s\in U}f(s)-2\epsilon$.
    By submodularity we have that $\kappa\max_{s\in U}f(s)\geq f(OPT)$. Therefore, $\Delta f(S,s)\geq \frac{f(OPT)-f(S)}{\kappa}-2\epsilon$.
    
    At the round $i$ in the for loop from Line \ref{line:tg_for_loop_starts} to Line \ref{line:tg_for_loop_ends} where $i>1$, if an element $o\in OPT$ is not added to the solution set, then it is not added to the solution at the last iteration, where the threshold is $\frac{w}{1-\alpha}$. By Lemma \ref{lem:clean_event}, we have
    $\Delta f(S,o)\leq \frac{w}{1-\alpha}+\epsilon$.
    Since for any element $s$ that is added to the solution at iteration $i$, by Lemma \ref{lem:clean_event} it holds that $\Delta f(S,s)\geq w-\epsilon$. Therefore, we have
    \begin{align*}
        \Delta f(S,s)&\geq w-\epsilon \\
        &\geq(1-\alpha)(\Delta f(S,o)-\epsilon)-\epsilon\\
        &\geq (1-\alpha)\Delta f(S,o)-2\epsilon. 
    \end{align*}
    By submodularity, it holds that $\Delta f(S,s)\geq (1-\alpha)\frac{f(OPT)-f(S)}{\kappa}-2\epsilon$.
\end{proof}
We now prove Theorem \ref{thm:threshold}, which relies on the previous Lemma \ref{lem:clean_event_monotone}, \ref{thm:samp} and \ref{lem:mar_gain}.
\begin{proof}
    The events defined in Lemma \ref{thm:samp}, \ref{lem:clean_event_monotone} hold with probability at least $1-\delta$ by combining Lemma \ref{thm:samp}, \ref{lem:clean_event_monotone}, and taking the union bound. Therefore in order to prove Theorem \ref{thm:threshold}, we assume that both the two events have occurred. 
    The proof of the first result in the theorem depends on the Lemma \ref{lem:mar_gain}. First, consider the case where the output solution set satisfies $|S|=\kappa$. Denote the solution set $S$ after the $i$-th element is added as $S_i$. Then by Lemma \ref{lem:mar_gain}, we have
    \begin{align*}
        f(S_{i+1})\geq \frac{1-\alpha}{\kappa}f(OPT)+(1-\frac{1-\alpha}{\kappa})f(S_i)-2\epsilon.
    \end{align*}
By induction, we have that
\begin{align*}
    f(S_{\kappa})&\geq(1-(1-\frac{1-\alpha}{\kappa})^{\kappa})\{f(OPT)-\frac{2\kappa\epsilon}{1-\alpha}\}\\
    &\geq (1-e^{-1+\alpha})\{f(OPT)-\frac{2\kappa\epsilon}{1-\alpha}\}\\
    &\geq(1-e^{-1}-\alpha)\{f(OPT)-\frac{2\kappa\epsilon}{1-\alpha}\}\\
    &\geq(1-e^{-1}-\alpha)f(OPT)-2\kappa\epsilon.
\end{align*}
 If the size of the output solution set $S$ is smaller than $\kappa$, then any element $o\in OPT$ that is not added to $S$ at the last iteration satisfies that
$\Delta f(S,o)\leq w+\epsilon$.
Since the threshold $w$ in the last iteration satisfies that $w\leq\frac{\alpha d}{\kappa}$, we have
\begin{align*}
    \Delta f(S,o)&\leq\frac{\alpha d}{\kappa}+\epsilon.
\end{align*}
It follows that
\begin{align*}
    \sum_{o\in OPT\backslash S}\Delta f(S,o)
    &\leq\alpha (\max_{s\in S}f(s)+\epsilon)+\kappa\epsilon\\
    &\leq\alpha f(OPT)+2\kappa\epsilon.
\end{align*}
 By submodularity and monotonicity of $f$, we have $f(S)\geq (1-\alpha)f(OPT)-2\kappa\epsilon$.

\end{proof}

    

\section{Appendix for Section \ref{sec:exp}}

\label{appdx:exp}
In this section, we present supplementary material to Section \ref{sec:exp}. In particular, we present the missing details about the experimental setup in Section \ref{appdx:setup}, and the additional experimental results in Section \ref{appdx:exp_results}.

\subsection{Experimental setup}
\label{appdx:setup}
First of all, we provide more details about the movie recommendation dataset MovieLens 25M dataset \cite{harper2015movielens} used to evaluate the algorithms proposed in the main paper. The dataset contains ratings on a 5-star scale of $X=13,816$ movies by the users. Each movie $x$ is associated with $1,128$ topics and a relevance score is assigned to each movie and each topic. Here we use $G(x,i)$ to denote the relevance score of movie $x$ and the topic $i$. In addition, we use $R(a,x)$ to denote the rating of user $a$ for the movie $x$. Since the number of topics is too large for our experiments, we select a small subset of topics by the following procedure. To extract the most relevant topics, we conduct correlation analysis to select a subset of topics. For each pair of features $i$ and $j$, if $Cor(i,j)\geq 0.4$ then either $i$ or $j$ is removed from the tag list. Subsequently, we remove any features that are negatively or weakly correlated to the ratings, i.e., $\forall i $ if $Cor(i, s)\leq 0.2$ then $i$ is removed from the list. In this way, we select $79$ tags. Then we randomly select different number of topics to use as the features.

In the experiments, the submodular basis functions $\{\vect{F}_i\}_{i\in[d]}$ is defined by the probabilistic coverage model. Next, we present the definition of the probabilistic coverage model, which captures how well a topic is covered by a subset of items \cite{hiranandani2020cascading,yu2016linear}. The problem
definition is as follows.
\begin{definition} (probabilistic coverage model)
    Suppose there are a total of $n$ elements denoted as $U$. For each element $x$ and each topic $i$, let $G(x,i)$ be the relevance score that quantifies how relevant the movie $x$ is to the topic $i$. For any tag $i$ and subset of elements $S$, function $F_i(S)$ is defined to be
\begin{align*}
    F_i(S)=1-\prod_{x\in S}(1-G(x,i)).
\end{align*}
Therefore, the marginal gain of adding a new movie $x$ to a subset $S$ is defined as 
\begin{align*}
    \Delta F_i(S,x)=G(x,i)\prod_{x'\in S}(1-G(x',i)),  \qquad\forall x\notin S.
\end{align*}
\end{definition}

For each user, we calculate the preference of the user for different topics as follows.
\begin{align*}
    w(a,i)=\frac{\sum_{x\in X}R(a,x)G(x,i)}{\sum_{x\in X,i'\in[d]}R(a,x)G(x,i')},
\end{align*}
where $X$ is the set of movies.

Next, we describe more details about the algorithms that we compare to in the experiments. The TG algorithm that we compare to is based on the threshold greedy algorithm with the evaluation of marginal gain being replaced by the repeated sampling approach. Therefore, by applying the Hoeffding's inequality, TG achieves the same approximation ratio as our algorithms. 
The algorithm EXP-GREEDY is proposed in \cite{singla2016noisy} and is based on the standard greedy algorithm with the element selecting procedure in each greedy round replaced by the CLUCB algorithm for best arm identification for general bandit. The algorithm TG doesn't utilize the linear structure either. As a result, these algorithms can't be used to reuse previous samples.

\subsection{Exprimental results}
\label{appdx:exp_results}

\begin{figure*}[t!]
    \centering
    \hspace{-0.5em}
     \subfigure[movie60 $\kappa$]
{\label{fig:movie60kf}\includegraphics[width=0.3\textwidth]{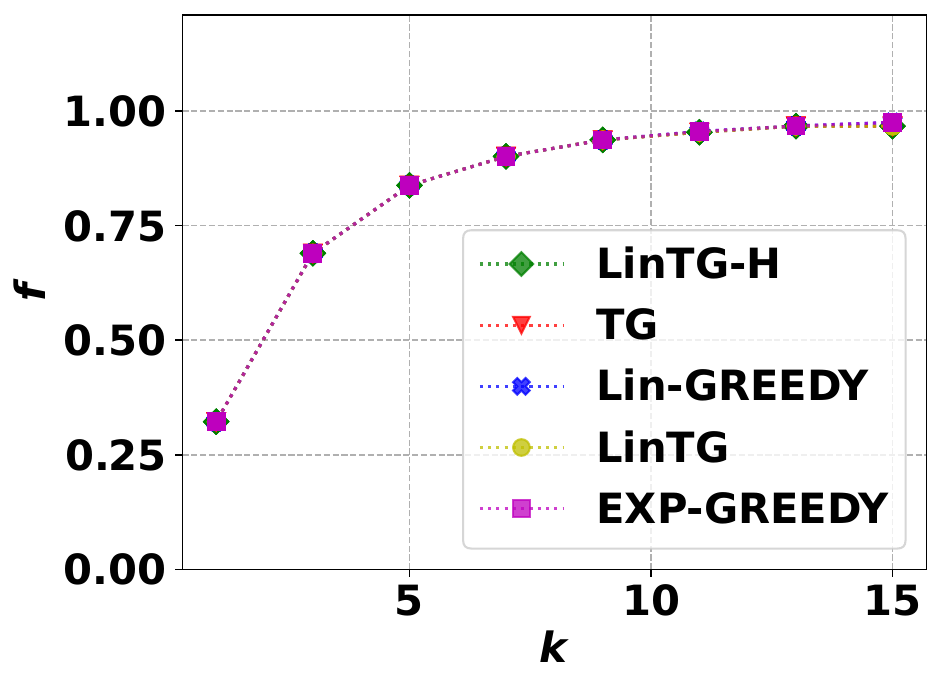}} 
\hspace{-0.5em}
     \subfigure[movie5000 $\kappa$]
{\label{fig:movie5000kf}\includegraphics[width=0.3\textwidth]{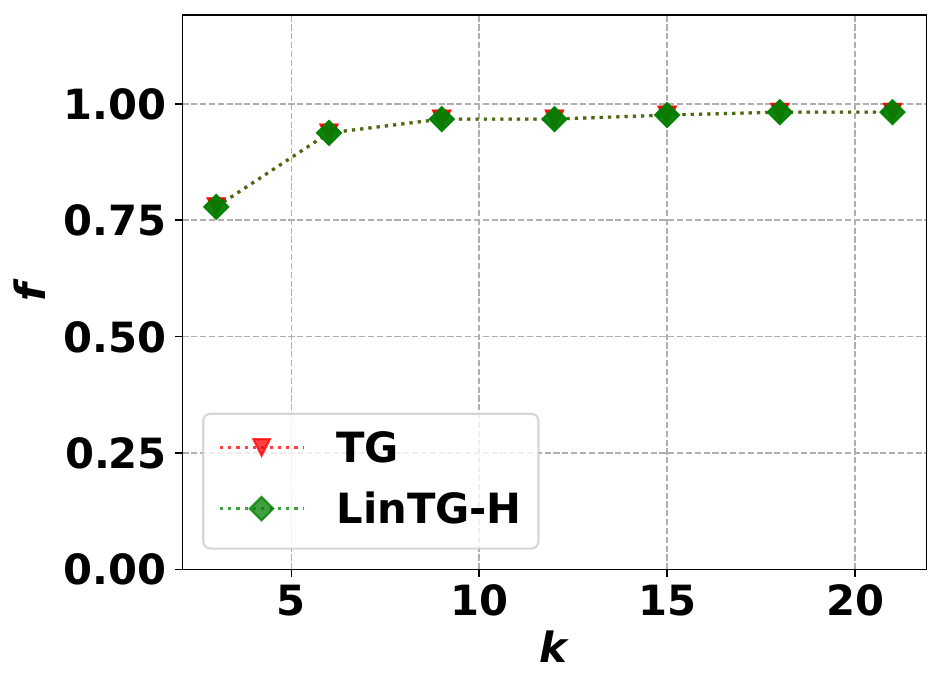}} 
    \hspace{-0.5em}
     \subfigure[movie60 $\epsilon$]
{\label{fig:movie60epsf}\includegraphics[width=0.3\textwidth]{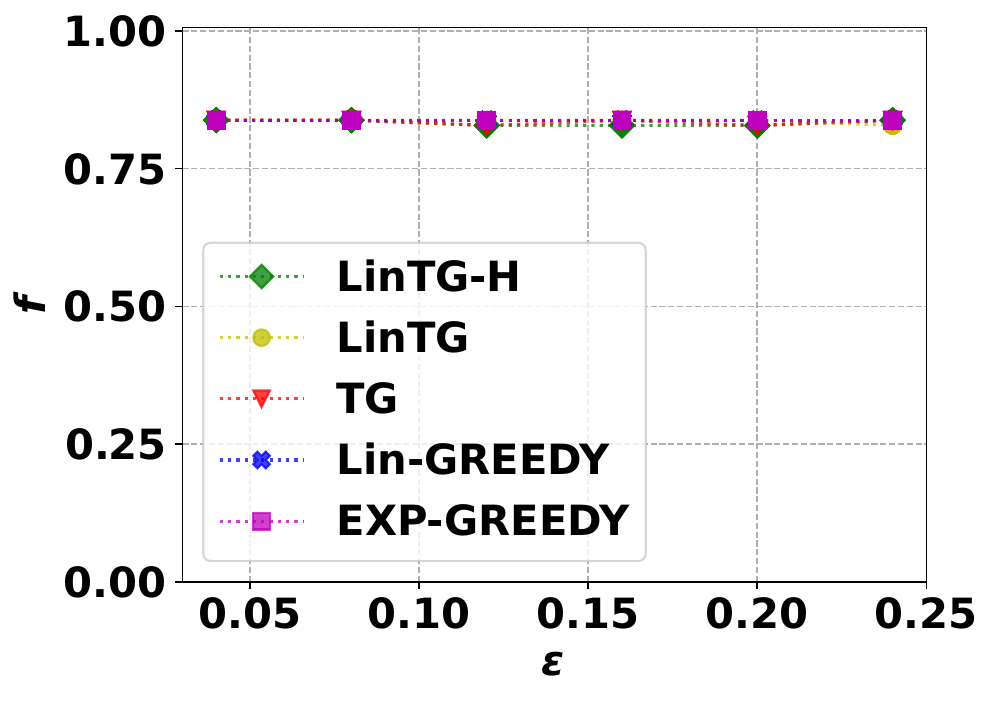}} 
\hspace{-0.5em}
\subfigure[movie500 d]
{\label{fig:movie500df}\includegraphics[width=0.3\textwidth]{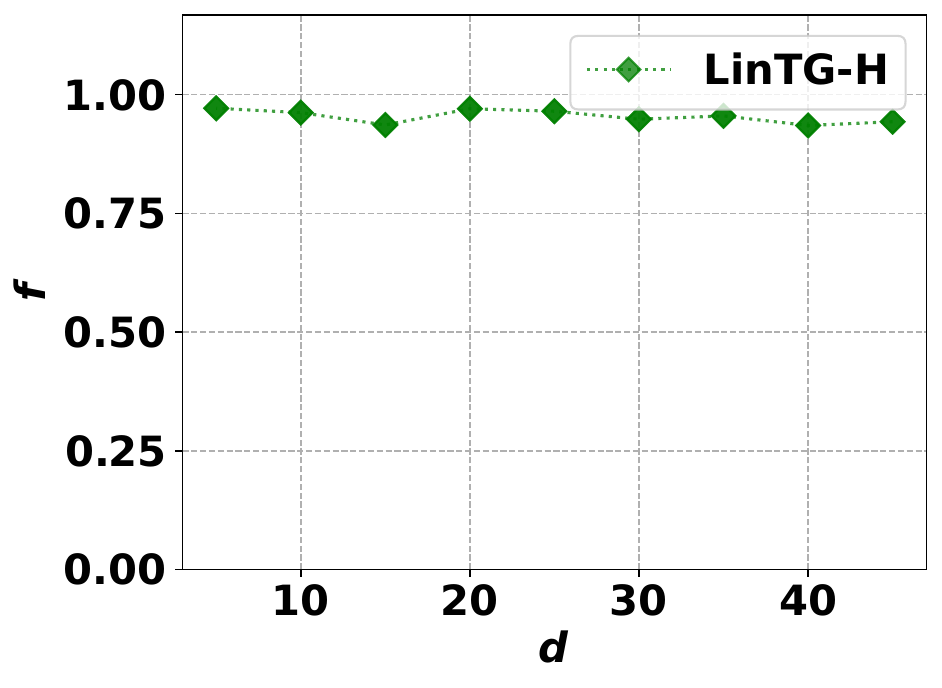}}
\hspace{-0.5em}
\subfigure[movie5000 $\epsilon$]
{\label{fig:movie5000epsf}\includegraphics[width=0.3\textwidth]{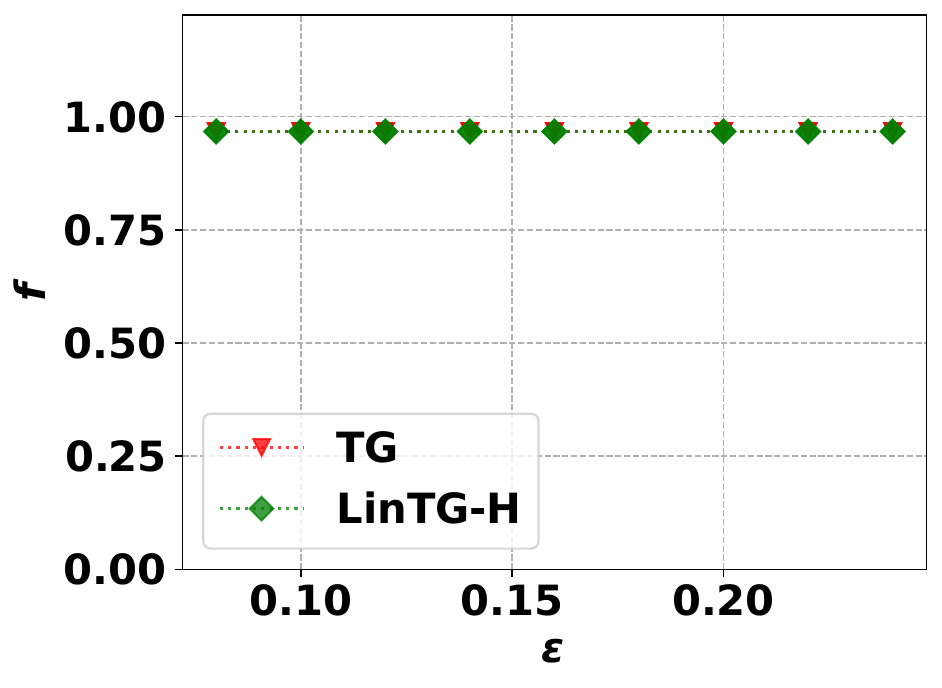}}
\hspace{-0.5em}
\subfigure[movie5000 $\epsilon$]
{\label{fig:movie5000epsq}\includegraphics[width=0.3\textwidth]{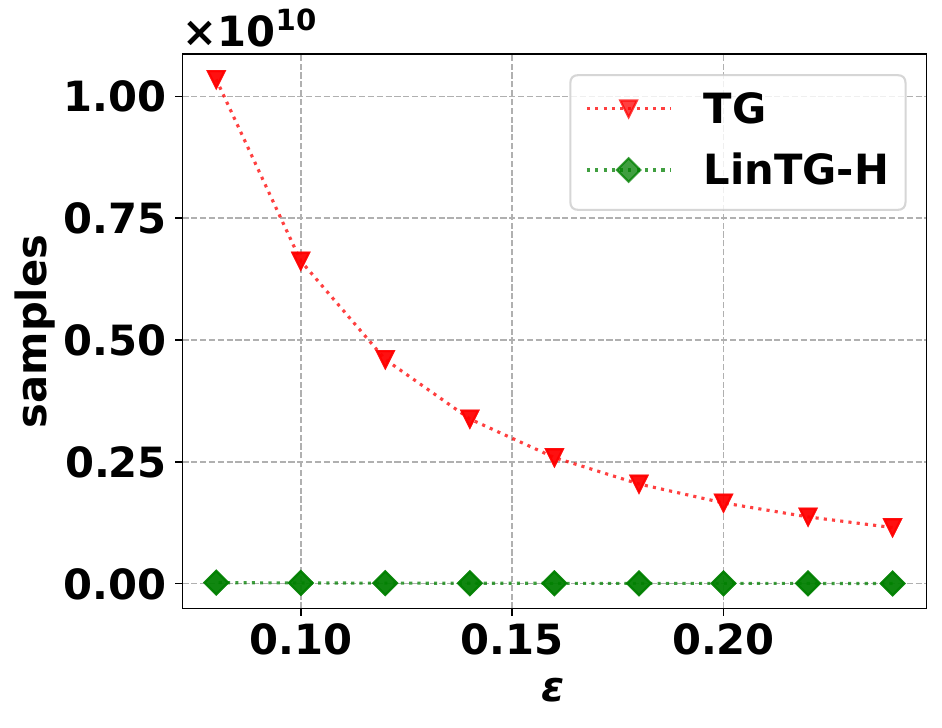}}
\hspace{-0.5em}
\caption{The experimental results of running the algorithms on instances of movie recommendation on the subsets of MovieLens 25M dataset with $n=60$, $d=5$ ("movie60") and $n=5000$, $d=30$ ("movien5000"), and different datasets with different value of $d$.}
        \label{fig:appdx}
\end{figure*}

The additional results of the experiments are presented in Figure \ref{fig:appdx}. From the results, we can see that the exact function value of different algorithms are almost the same. On the "movie5000" dataset, the sample complexity of our proposed method LinTG-H is much smaller than TG, which demonstrates the sample efficiency of our algorithm. 

\section{Technical Lemmas and Details}

 \subsection{ Matrix Computations and Runtime}
 \label{appdx:runtime}
 In this section, we present the missing discussions on the implementation and runtime of the matrix inverse and multiplications in the main paper.

\textbf{Inverse of matrix. } Notice that all of the subroutine algorithms presented in the main paper keep track of $\vect{A}_t^{-1}$  throughout the algorithm. From Lemma \ref{lem:matrix_inversion},  
 we can see that the inverse of $(\vect{A}_{t}+\vect{x}\vect{x}^T)$  can be computed as 
$$(\vect{A}_{t}+\vect{x}\vect{x}^T)^{-1}=\vect{A}_{t}^{-1}-\frac{(\vect{A}_{t}^{-1}\vect{x})(\vect{A}_{t}^{-1}\vect{x})^T}{1+\vect{x}^T\vect{A}_{t}^{-1}\vect{x}},$$
 for any $\vect{x}\in U$.
 This step takes $O(d^2)$ in time complexity.

 \subsection{Technical Lemmas}
 \label{appdx:tech_lem}

\begin{proposition}
\label{prop:staticy} (Proposition 1 in \cite{soare2014best}). Suppose the feature vector set is denoted as $\mathcal{X}$ with $|\mathcal{X}|=n$.
    Let $\hat{\vect{w}}_t$ 
be the solution to the least-squares problem and let $\vect{A}_t
 = \vect{X}_t^T\vect{X}_t$, and assume the noise variable $\xi_t$ is $R$-sub-Gaussian. Assume $\mathcal{Y}$ is a subset of $\mathbb{R}^d$ with that $|\mathcal{Y}|=Y$.  Then for all $\vect{y}\in\mathcal{Y}$ and any fixed sequence $\vect{x}_t$, we have that with probability at least $1-\delta$, it holds that
\begin{align*}
|\vect{y}^T(\hat{\vect{w}}_t-\vect{w})|\leq R||\vect{y}||_{(\vect{A}_t)^{-1}}\sqrt{2\log(\frac{\pi^2t^2Y}{3\delta})}
\end{align*}
for all $t>0$.

\end{proposition}
 
\textbf{Proposition \ref{prop:adaptive}. }\textit{
    Let $\hat{\vect{w}}_t^{\lambda}$ 
be the solution to the regularized least-squares problem with
regularizer $\lambda$ and let $\vect{A}_t^{\lambda}
 = \vect{X}_t^T\vect{X}_t+\lambda \vect{I}$. Then for every adaptive sequence $\vect{X}_t$ such
that at any step t, $\vect{x}_{a_t}$ only depends on past history, we have that with probability at least $1-\delta$, it holds that for all $t\geq 0$ and all $\vect{y}_t\in \mR^d$ that only depends on past history up to time $t$,
\begin{align*}
|\vect{y}_t^T(\hat{\vect{w}}_t^{\lambda}-\vect{w})|\leq ||\vect{y}_t||_{(\vect{A}_t^{\lambda})^{-1}}C_t, 
\end{align*}
 where $C_t$ is defined as 
\begin{align*}
C_t=R\sqrt{2\log\frac{\det(\vect{A}_t^{\lambda})^{\frac{1}{2}}\det(\lambda I)^{-\frac{1}{2}}}{\delta}}+\lambda^{\frac{1}{2}}S.
\end{align*}
Moreover, if $||\vect{x}_{a_t}||\leq L$ holds for all $t > 0$, then
\begin{align*}
    C_t\leq R\sqrt{d\log(\frac{1+tL^2/\lambda}{\delta})}+\lambda^{\frac{1}{2}}S.
\end{align*}
}

\begin{proof}
    Let us denote the history up to time t as $\mathcal{H}_t$, then by Cauchy–Schwarz inequality, we have that for each fixed time $t$
    \begin{align*} P(|\vect{y}_t^T(\hat{\vect{w}}_t^{\lambda}-\vect{w})|\leq ||\vect{y}||_{(\vect{A}_t^{\lambda})^{-1}}C_t, \forall \vect{y}_t \text{ depend on } \mathcal{H}_t|\mathcal{H}_t)\geq P(||\hat{\vect{w}}_t^{\lambda}-\vect{w}||_{(\vect{A}_t^{\lambda})^{-1}}\leq C_t|\mathcal{H}_t) 
    \end{align*}
    It then follows that 
     \begin{align*} P(|\vect{y}_t^T(\hat{\vect{w}}_t^{\lambda}-\vect{w})|\leq ||\vect{y}||_{(\vect{A}_t^{\lambda})^{-1}}C_t, \forall \vect{y}_t \text{ depend on } \mathcal{H}_t)\geq P(||\hat{\vect{w}}_t^{\lambda}-\vect{w}||_{(\vect{A}_t^{\lambda})^{-1}}\leq C_t). 
    \end{align*}
    From Theorem 2 in \cite{abbasi2011improved}, we can see that with probability at least $1-\delta$, at any step $t$,
    $||\hat{\vect{w}}_t^{\lambda}-\vect{w}||_{(\vect{A}_t^{\lambda})^{-1}}\leq C_t$. Therefore, by applying this result and taking a union bound over all time step $t$, we can prove the result in the proposition.
\end{proof}

    \begin{lemma} \label{lem:matrix_inversion}
        (Woodbury formula. ) The inverse of matrix $A+UCV$ can be computed as
        \begin{align*}
            (A+UCV)^{-1}=A^{-1}-A^{-1}U(C^{-1}+VA^{-1}U)^{-1}VA^{-1}.
        \end{align*}
    \end{lemma}
\begin{lemma}
    \label{lem:bound_det_A}
    (\cite{abbasi2011improved}, Lemma
10) Let the maximum $l_2$-norm of the feature vectors be denoted
as $L$. The determinant of matrix $\vect{A}_t$ is bounded as
\begin{align*}
    det(\vect{A}_t)\leq(\lambda+tL^2/d)^d
\end{align*}
\end{lemma}

\begin{lemma}
\label{lem:opt_equal}
For any vector $\vect{x}_1,\vect{x}_2...\vect{x}_m\in\mathbb{R}^d$ and  $\vect{y}\in \mathbb{R}^d$ and , if we set $p_a^*$ to be 
\begin{align*}
    p_a^*=\frac{|w^*_a|}{\sum_{a'=1}^m|w^*_{a'}|},
\end{align*}
 where $w^*_{a'} $ is the optimal solution of the linear program in (\ref{eqn:optimization}) below 
\begin{align}
\label{eqn:opt_prob}
    \arg\min_{\{w_a\}}\sum_{a=1}^m|w_a| \qquad s.t.\qquad \vect{y}=\sum_{a=1}^mw_a\vect{x}_a.
\end{align}
Then $p_a^*$ is also the asymptotic optimal solution of the optimization problem below when $n\rightarrow\infty$. 
\begin{align*}
    \arg\min_{\{p_a\}}~&\vect{y}^T(\sum_{a\in[m]}p_a\vect{x}_a\vect{x}_a^T+\frac{\lambda}{n}I)^{-1}\vect{y}\\
    s.t. ~&\sum_{a\in[m]}p_a=1.
\end{align*}
Moreover, we have that
\begin{align*}
    \sqrt{\vect{y}^T(\sum_{a\in[m]}p_a^*\vect{x}_a\vect{x}_a^T)^{-1}\vect{y}}\leq \rho,
\end{align*}
where $\rho$ is the optimal value of the optimization problem in (\ref{eqn:opt_prob}).
\end{lemma}
\begin{proof}
    The proof can be found in Appendix B in \cite{xu2018fully}, so we omit the proof here.
\end{proof}

    \begin{lemma}
    \label{lem:logx_over_x}
    Suppose $x\in\mathbb{R}$ and $x\geq 2$, if we have $x\geq\frac{2}{a}\log\frac{2}{a}$, then it holds that 
    \begin{align*}
        \frac{\log x}{x}\leq a
    \end{align*}
\end{lemma}
\begin{proof}
    Since $y=\frac{\log x}{x}$ is decreasing when $x\geq 2$, if $x>\frac{2}{a}\log\frac{2}{a}$, then we have 
    \begin{align*}
        \frac{\log x}{x}< \frac{a}{2} \cdot \frac{\log (\frac{2}{a}\log\frac{2}{a})}{\log \frac{2}{a}}\leq a.
    \end{align*}
\end{proof}

\begin{lemma}
    \label{lem:calculation1}
    Suppose $x, a,b\in\mathbb{R}_+$, if we have $x\leq a+b\log x$, then it holds that 
    \begin{align*}
        x\leq 2a+2b\log b.
    \end{align*}
\end{lemma}




\end{document}